\documentclass[10pt,twocolumn,letterpaper]{article}

\usepackage{iccv}
\usepackage{times}
\usepackage{epsfig}
\usepackage{graphicx}
\usepackage{bm}
\usepackage{mathcommand}
\usepackage{mathrsfs}
\usepackage{amsmath,amsthm,amssymb,mathtools}
\usepackage{subfigure}
\usepackage{times}
\usepackage{epsfig}
\usepackage{graphicx}
\usepackage{amsmath}
\usepackage{amssymb}
\usepackage{bm}
\usepackage{amsthm}
\usepackage{algorithm}
\usepackage{algorithmicx}
\usepackage{algpseudocode}
\usepackage[dvipsnames]{xcolor}
\usepackage{caption}
\usepackage[export]{adjustbox}
\usepackage{float}
\usepackage{stfloats}
\usepackage{mathtools}
\usepackage{tabularx}
\usepackage{booktabs}
\usepackage{float}
\usepackage{placeins}
\usepackage{thm-restate}
\usepackage{thmtools}
\usepackage{wrapfig}
\usepackage{multirow}
\usepackage{enumitem}
\setenumerate[1]{itemsep=0pt,partopsep=0pt,parsep=\parskip,topsep=5pt}
\setitemize[1]{itemsep=0pt,partopsep=0pt,parsep=\parskip,topsep=5pt}
\setdescription{itemsep=0pt,partopsep=0pt,parsep=\parskip,topsep=5pt}
\newcommand{\bfq}{\mathbf{q}}
\newcommand{\bfk}{\mathbf{k}}
\newcommand{\bfv}{\mathbf{v}}
\newcommand{\bfx}{\mathbf{x}}

\newcommand{\bfz}{\mathbf{z}}
\newcommand{\bfA}{\mathbf{A}}

\newcommand{\bfW}{\mathbf{W}}

\newcommand{\bfw}{\mathbf{w}}

\newcommand{\bfM}{\mathbf{M}}

\newcommand{\bfI}{\mathbf{I}}

\newcommand{\bff}{\mathbf{f}}

\newcommand*{\tran}{^{\mkern-1.5mu\mathsf{T}}}

\def\aka{a.k.a\onedot}

\newcommand{\ud}{\mathrm{d}}

\usepackage[pagebackref=true,breaklinks=true,letterpaper=true,colorlinks,bookmarks=true]{hyperref}

\iccvfinalcopy 

\def\iccvPaperID{11210} 

\ificcvfinal\fi

\begin{document}

\title{TF-ICON: Diffusion-Based Training-Free Cross-Domain Image Composition}

\author{
{Shilin Lu}\textsuperscript{1} \quad {Yanzhu Liu}\textsuperscript{2} \quad {Adams Wai-Kin Kong}\textsuperscript{1}\\
\textsuperscript{1}School of Computer Science and Engineering, Nanyang Technological University, Singapore \\ \hspace{-0.4cm}\textsuperscript{2}Institute for Infocomm Research ($\rm{I^2R}$) \& Centre for Frontier AI Research (CFAR), A*STAR, Singapore \\
\tt\small shilin002@e.ntu.edu.sg,
liu\_yanzhu@i2r.a-star.edu.sg, \tt\small adamskong@ntu.edu.sg\\
}

\twocolumn[{
	\renewcommand\twocolumn[1][]{#1}
	\maketitle
	\centering
	\vspace*{-1cm}
	\includegraphics[width=1\textwidth]{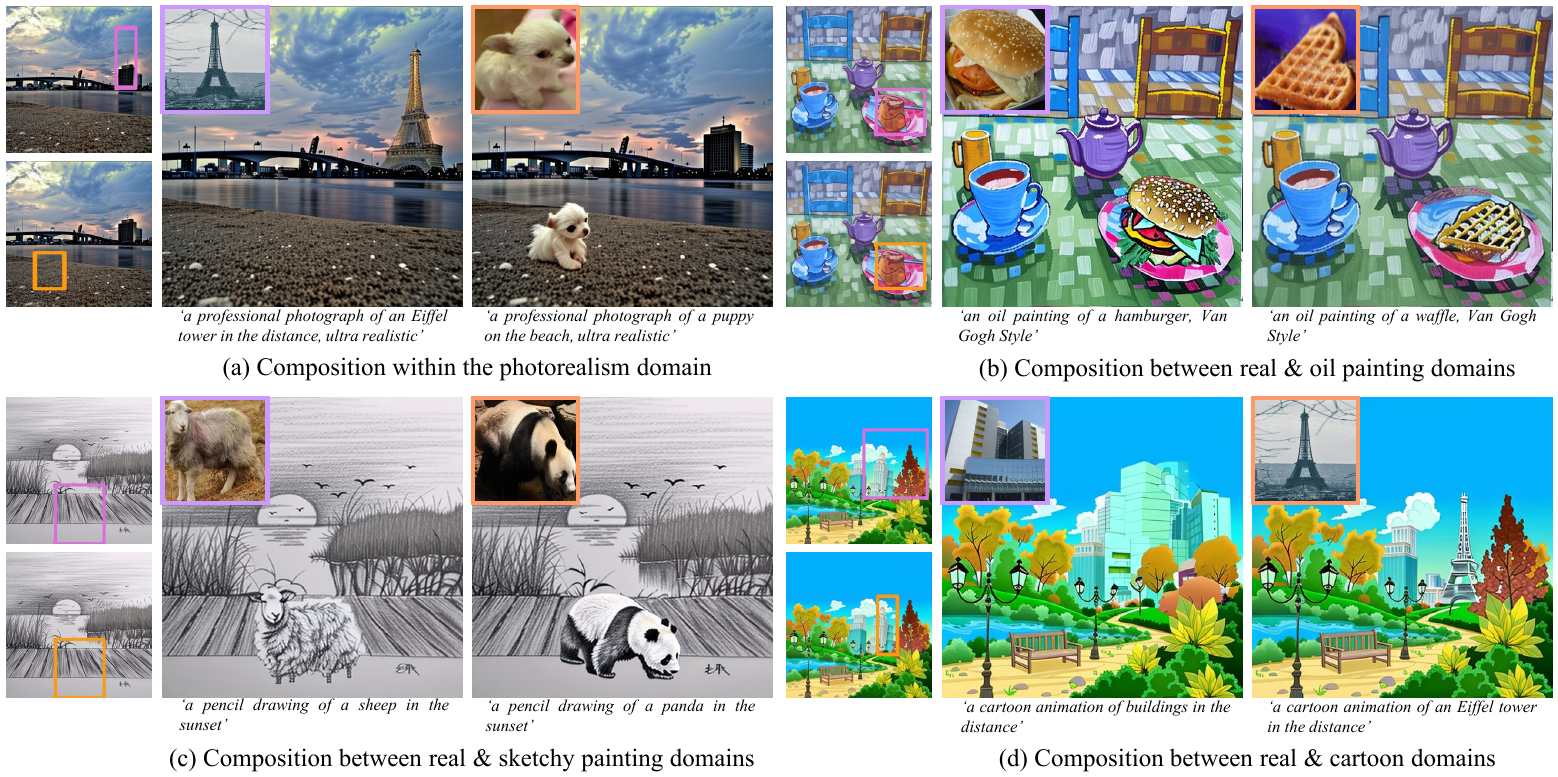} 
	\vspace{-0.6cm}
	\captionof{figure}{Image composition aims to seamlessly blend distinct objects into a specific visual context. Our training-free framework equips attention-based text-driven diffusion models with the capability to achieve this task across various domains (a) photorealism, (b) oil painting, (c) sketching, and (d) cartoon animation, within 20 sampling steps.}
	\label{fig:our_results}
	\vspace*{0.3cm}
}]

\ificcvfinal\thispagestyle{empty}\fi

\hypersetup{
	bookmarksnumbered=true,
}

\begin{abstract}
\vspace{-0.4cm}

Text-driven diffusion models have exhibited impressive generative capabilities, enabling various image editing tasks. In this paper, we propose TF-ICON, a novel Training-Free Image COmpositioN framework that harnesses the power of text-driven diffusion models for cross-domain image-guided composition. This task aims to seamlessly integrate user-provided objects into a specific visual context. Current diffusion-based methods often involve costly instance-based optimization or finetuning of pretrained models on customized datasets, which can potentially undermine their rich prior. In contrast, TF-ICON can leverage off-the-shelf diffusion models to perform cross-domain image-guided composition without requiring additional training, finetuning, or optimization. Moreover, we introduce the exceptional prompt, which contains no information, to facilitate text-driven diffusion models in accurately inverting real images into latent representations, forming the basis for compositing. Our experiments show that equipping Stable Diffusion with the exceptional prompt outperforms state-of-the-art inversion methods on various datasets (CelebA-HQ, COCO, and ImageNet), and that TF-ICON surpasses prior baselines in versatile visual domains. Code is available at \href{https://github.com/Shilin-LU/TF-ICON}{https://github.com/Shilin-LU/TF-ICON}
\end{abstract}

\vspace{-0.6cm}
\section{Introduction}
Image composition task involves incorporating unique objects from different photos to create a harmonious image within a specific visual context, \aka image-guided composition. For instance, consider the scenario where one desires to incorporate a beloved panda into one’s favorite artwork, \eg, oil or sketchy painting. The objective is to create a new image where the panda blends seamlessly into the scene without altering the appearance of the panda and the background, just as an artist meticulously crafted this panda for that artwork (See Figure~\ref{fig:our_results}). This task is inherently challenging, as it requires maintaining illumination consistency and preserving identifying features. The challenge is further compounded when the photos come from various domains. 

While recently large-scale text-to-image models \cite{chang2023muse, ding2022cogview2, nichol2021glide, ramesh2022hierarchical, rombach2022high, saharia2022photorealistic,  yu2022scaling} have achieved remarkable success in text-driven image generation, the ambiguity inherent in natural language presents challenges in conveying precise and nuanced visual details, even with highly detailed text prompts. Although this challenge is effectively addressed by enabling personalized concept learning \cite{gal2022image, gal2023designing, kawar2022imagic, kumari2022multi, ruiz2022dreambooth}, these methods require costly instance-based optimization and are limited in generating concepts with specified backgrounds. Recent studies \cite{song2022objectstitch, yang2022paint} have shown that diffusion models can achieve image-guided composition by explicitly incorporating additional guiding images. However, these models are retrained from the pretrained diffusion model on tailored datasets, which can damage the rich prior of the model. As a result, these models have limited compositional abilities beyond their training domain and still require significant computational resources.


Given the wealth of large text-to-image models that have been trained on extensive language-image datasets, we pose a question: \textit{how could these models be leveraged for image-guided composition without incurring costly training or finetuning, thereby avoiding damaging the diverse prior?} To answer it, we propose the Training-Free Image COmpositioN (TF-ICON) framework, which equips attention-based text-to-image diffusion models with the capability to perform image-guided composition without requiring additional training, fine-tuning, extra data, or optimization. To the best of our knowledge, this is the first training-free framework developed for image-guided composition. The framework is compatible with various diffusion model samplers, enabling completion within 20 steps, and harnesses rich semantic knowledge to facilitate image-guided compositions across diverse domains (see Figure~\ref{fig:our_results}). 

Our approach constitutes an image-guided composition interface through denoising from a reliable starting latent code with the injection of composite self-attention maps. Finding the latent code that allows for reconstructing an input image while maintaining its editability, \aka image inversion, is a challenging yet crucial step for state-of-the-art (SOTA) image editing frameworks involving real images \cite{couairon2022diffedit, hertz2022prompt, kim2022diffusionclip, kwon2022diffusionb, meng2021sdedit, parmar2023zero, patashnik2021styleclip, tumanyan2022plug}. For diffusion models, while denoising diffusion implicit models (DDIM) inversion \cite{song2020denoising} has been effective for unconditional diffusion models, it falls short for text-driven diffusion models \cite{hertz2022prompt, mokady2022null, tumanyan2022plug, wallace2022edict}. To circumvent this, we introduce the exceptional prompt to accurately invert real images into latent codes upon pretrained text-to-image models to serve for further composition generation. The accurate latent codes are composed as the starting noise for the diffusion process. Through the gradual injection of composite self-attention maps that are specifically designed to reflect the relations between guiding images, we are able to infuse contextual information from the background into the incorporated objects, which results in harmonious image-guided compositions. 

To summarize, we make the following key contributions:
\begin{itemize}
	\setlength{\itemsep}{0pt}
	\setlength{\parsep}{0pt}
	\setlength{\parskip}{0pt}
	\item [1.] We demonstrate the superior performance of high-order diffusion ODE solvers compared to commonly used DDIM inversion for real image inversion.
	\item [2.] We present an exceptional prompt that allows text-driven models to achieve accurate invertibility, laying a solid groundwork for subsequent editing. Experimental results show that it surpasses SOTA inversion methods on three vision datasets.
	\item [3.] We propose the first training-free framework that enables cross-domain image-guided composition for attention-based diffusion models.
	\item [4.] We demonstrate quantitatively and qualitatively that our framework outperforms prior baselines for image-guided composition.
\end{itemize}

\section{Related Work}
\noindent \textbf{Image composition.} Image composition is widely applied to electronic commerce, entertainment, and data augmentation \cite{dwibedi2017cut,liu2021opa} for downstream tasks. It can be broadly categorized into two types: text-guided \cite{avrahami2022blended_latent, avrahami2022blended, chefer2023attend, feng2022training, liu2022compositional} and image-guided \cite{brown2022end, gafni2020wish, li2023gligen, song2022objectstitch, xue2022dccf, yang2022paint, zhang2021deep}. The former involves composing multiple objects specified by only a text prompt without limiting the appearance of objects, as long as their semantics align with the prompt. Despite the great successes of text-conditioned models, they are often prone to semantic errors \cite{feng2022training, ramesh2022hierarchical}, especially when the text prompt involves multiple objects. These errors include attribute leakage, attribute interchange, and missing objects, which cause the generated images to critically different from the user's intention \cite{feng2022training, ramesh2022hierarchical}. As a result, extensive prompt engineering \cite{witteveen2022investigating} is often necessary to achieve the desired results. In contrast to text-only guided composition, image-guided composition involves incorporating specific objects and scenarios from user-provided photos, potentially with the aid of a text prompt. However, the inclusion of additional real images poses a greater challenge, especially when merging images from different visual domains, which is the primary focus of painterly image harmonization \cite{cao2023painterly,lu2023painterly,zhang2020deep}. Conventionally, image-guided composition is divided into several sub-tasks \cite{niu2021making}, such as object placement \cite{azadi2020compositional, chen2019toward, lin2018st, tripathi2019learning, zhang2020learning}, image blending \cite{wu2019gp, zhang2020deep}, image harmonization \cite{cong2020dovenet, cun2020improving, jiang2021ssh, xue2022dccf}, and shadow generation \cite{hong2022shadow, liu2020arshadowgan, sheng2021ssn, zhang2019shadowgan}, each of which is typically addressed by different models and pipelines. \\

\noindent \textbf{Image inversion.} Extensive research has been conducted on image inversion for GANs, including latent-based optimization \cite{abdal2019image2stylegan, abdal2020image2stylegan, karras2020analyzing}, encoders \cite{alaluf2021restyle, richardson2021encoding, tov2021designing}, and fine-tuning \cite{alaluf2022hyperstyle, roich2022pivotal}. For diffusion models, DDIM \cite{song2020denoising} is a widely used technique for inversion in image editing frameworks. However, in text-driven settings, DDIM leads to significant reconstruction distortion due to the instability resulting from classifier-free guidance (CFG) \cite{dhariwal2021diffusion, ho2022classifier}. Recently, null-text inversion \cite{mokady2022null} has been proposed to achieve accurate inversion by optimizing the unconditional prediction of the text-to-image model. It demonstrates promising results but requires instance-based optimization. Concurrently, EDICT \cite{wallace2022edict} also achieves near-perfect inversion, albeit doubling the computation time of the diffusion process.

\begin{figure*}[htbp]
	\centering
	\vspace{-0.2cm}
	\includegraphics[width=1\linewidth]{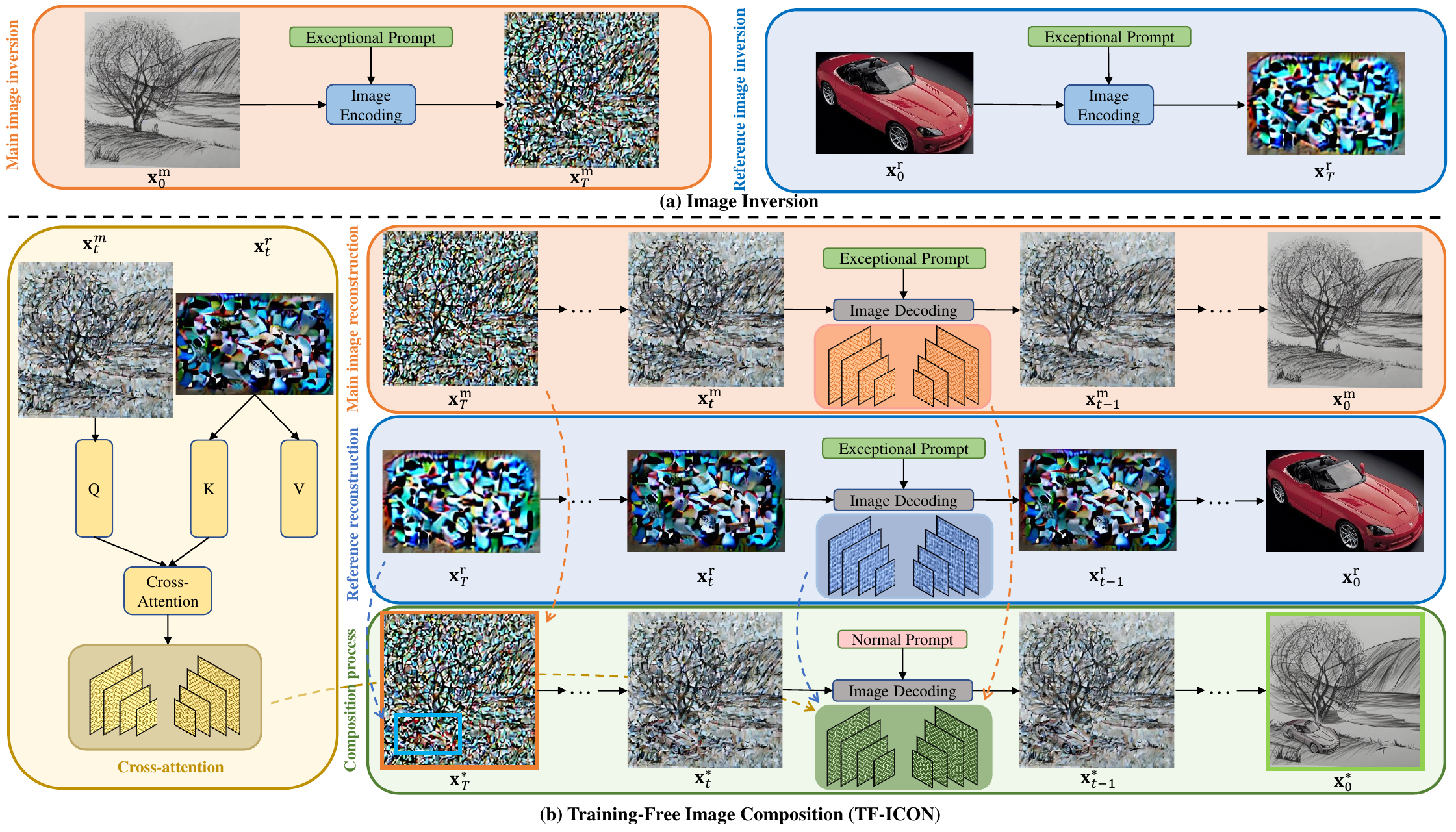}
	\vspace{-0.5cm}
	\caption{The proposed training-free image composition framework. (a) The exceptional prompt is used to invert the \textcolor[RGB]{255,128,0}{main} and \textcolor[RGB]{65,105,255}{reference} images into noises $\bfx^\text{m}_T, \bfx^\text{r}_T$, which are then composed to form the starting point $\bfx^*_T$ for the \textcolor[RGB]{0,201,87}{composition process}. (b) Three constituents are composed for injecting into the \textcolor[RGB]{0,201,87}{composition process} at early timesteps, including self-attention maps from the \textcolor[RGB]{255,128,0}{main} and \textcolor[RGB]{65,105,255}{reference} image reconstruction processes, as well as \textcolor[RGB]{156,102,31}{cross attention} between the main and reference images. For better clarity and readability, the original main and reference images are shown in the pixel space instead of the VAE latent space, and the reference image is presented without resizing and zero-padding.}
	\vspace{-0.2cm}
	\label{fig:overview}
\end{figure*}


\section{Preliminary}
\label{sec:sde}
Diffusion probabilistic models (DPM) \cite{dhariwal2021diffusion, ho2020denoising, rombach2022high, sohl2015deep} are generative models in which an image is generated by progressively denoising from Gaussian noise. The forward diffusion process gradually perturbs data with infinite noise scales, which can be modeled as the solution of a stochastic differential equation (SDE) $\{\bfx_t\}_{t=0}^T$. Formally, given a data sample $\bfx_0 \sim p_{0}=p_\mathrm{data}$, random noise is gradually injected, eventually resulting in a sample $\bfx_T$ which is typically distributed as a tractable prior $p_{T}$ without any information of $p_0$, as described by the following SDE \cite{kingma2021variational,song2020score}:

\begin{equation}
	\ud \bfx_t = \bff(\bfx_t, t)\ud t + g(t) \ud \bfw_t,
	\label{eqn:forward_sde}
\end{equation}
where $\bfw_t \in \mathbb{R}^d$ is the standard Wiener process (\aka, Brownian motion), and $\bff(\cdot, t)$ and $g(t)$ are commonly designated as the drift and diffusion coefficient, respectively. On the other hand, the reverse diffusion process can be described by the reverse-time SDE from $T$ to 0 \cite{song2020score}:
\begin{align}
	\ud \bfx_t = [\bff(\bfx_t, t) - g(t)^2  \nabla_{\bfx} \log p_t(\bfx_t)] \ud t + g(t) \ud \bar{\bfw}_t,\label{eqn:backward_sde}
\end{align}
where $\bar{\bfw}_t$ is the Wiener process in the reverse time. The score function $\nabla_{\bfx}  \log p_t(\bfx_t)$ is the only unknown term and can be estimated by a neural network $\epsilon_\theta(\bfx_t,t)$ whose parameter $\theta$ is optimized by a denoising objective \cite{ho2020denoising, song2020score}.



Upon attaining the trained model that predicts the score function accurately, it can be utilized to numerically solve the reverse SDE (Eq.~(\ref{eqn:backward_sde})), enabling the generation of samples from a noise distribution. Song \etal \cite{song2020score} outline various methods, including Variance Exploding (VE), Variance Preserving (VP), and sub-VP SDE, for constructing SDEs that perturb the unknown data distribution into a fixed prior. In this work, we leverage the pre-trained text-to-image Latent Diffusion Model (LDM) \cite{rombach2022high}, \aka Stable Diffusion, which applies the VP SDE in the latent space.

\section{Method}
Our objective is to utilize a main (background) image $\bfI^\text{m}$, a reference (foreground) image $\bfI^\text{r}$, a text prompt $\mathcal{P}$, and a binary mask $\bfM^\text{user}$ which designates the region of interest within the main image, to generate a modified image $\bfI^*$. The resultant image $\bfI^*$ should contain the reference subject with identifying features within the mask, \ie $id(\bfI^* \odot \bfM^\text{user})  \approx id(\bfI^\text{r})$, while concurrently ensuring that the complementing area closely resembles the main image, \ie $\bfI^* \odot (\bf{1}-\bfM^\text{user})  \approx \bfI^\text{m} \odot (\bf{1}-\bfM^\text{user})$. Moreover, it is ideal for the transition between the areas inside and outside the mask to be imperceptible. 

We propose a training-free framework that can make use of attention-based pre-trained text-to-image models to perform image-guided composition. To the best of our knowledge, it is the first training-free framework for image-guided composition, which can be accomplished within 20 steps of sampling. The framework is mainly comprised of two steps: \textbf{image inversion} (Section~\ref{sec:exceptional prompt}), and \textbf{composition generation} (Section~\ref{sec:atten-comp}), as shown in Figure~\ref{fig:overview}. The full algorithm is presented in Appendix \ref{sec:algo}.

\subsection{Image Inversion with Exceptional Prompt}
\label{sec:exceptional prompt}
Achieving precise manipulation of real images often necessitates an accurate inversion process that identifies the corresponding latent representation, which not only provides editability for meaningful manipulation but also accurately reconstructs the input image \cite{hertz2022prompt, tov2021designing}. For diffusion models, optimal editability is typically characterized by a noise encoding that conforms to the ideal statistical properties of zero-mean, unit-variance Gaussian noise \cite{parmar2023zero}. \\

\noindent \textbf{ODE inversion.} Most diffusion frameworks for image editing \cite{couairon2022diffedit, hertz2022prompt, kim2022diffusionclip, kwon2022diffusionb, parmar2023zero, tumanyan2022plug} use DDIM inversion to invert the real image into its latent representation. However, our findings suggest that this may not be the optimal choice for inverting real images. It has been proven that DDIM is a first-order discretization of the associated probability flow ordinary differential equations (ODE) of Eq.~(\ref{eqn:backward_sde}) \cite{salimans2022progressive, song2020denoising, song2020score}:
\begin{align}
	\ud \bfx_t = \Big[\bff(\bfx_t, t) - \frac{1}{2} g(t)^2\nabla_\bfx \log p_t(\bfx_t)\Big] \ud t, \label{eqn:flow}
\end{align}
which can be solved using $\epsilon_\theta(\bfx_t,t)$ and shares the consistent marginal probability distribution $\{p_t(\bfx_t)\}_{t=0}^T$ with Eq.~(\ref{eqn:backward_sde}). Various samplers \cite{karras2022elucidating, liu2022pseudo, lu2022dpma, lu2022dpmb} have been developed for solving the diffusion ODE starting from noise $\bfx_T$ to achieve fast sampling (10$\sim$20 steps). We offer the insight that utilizing these ODE solvers in turn as encoders starting from the real image $\bfx_0$ yields better latent representation $\bfx_T$, compared with those obtained through commonly used DDIM. A quantitative analysis is given in Appendix \ref{sec:better_latent_DPM}. The enhanced alignment between the forward and backward ODE trajectories in the high-order DPM-Solver++ \cite{lu2022dpmb} implies that it is better suited for real image inversion. Thus, this paper employs it for all inversions of diffusion models.



\begin{figure}[tbp]
	\centering
	\includegraphics[width=1\linewidth]{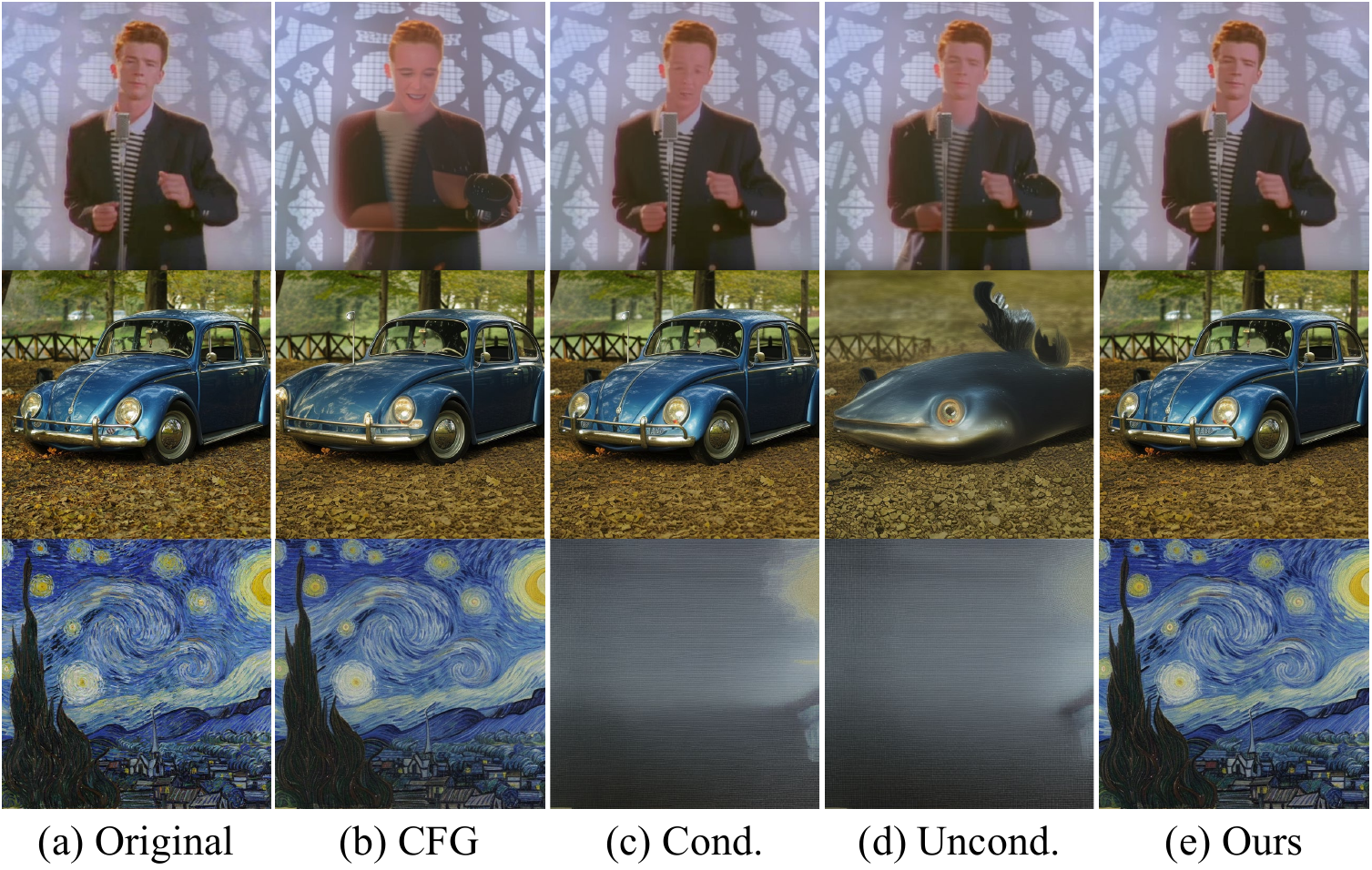}
	\vspace{-0.5cm}
	\caption{The real image reconstruction results using Stable Diffusion with (b) classifier-free guidance (CFG) output $\hat{\epsilon}_\theta(\bfx_t, t, \mathcal{E}, \varnothing)$; (c) conditional output $\epsilon_\theta(\bfx_t, t, \mathcal{E})$; (d) unconditional output $\epsilon_\theta(\bfx_t, t, \varnothing)$; and (e) ours. The prompts for (b) and (c) are \textit{`a photo of a singer'}, \textit{`a photo of a car'}, and \textit{`an oil painting'}. See Appendix \ref{sec:elab_for_fig3} for elaboration.}
	\vspace{-0.2cm}
	\label{fig:inversion}
\end{figure}

~\\
\noindent \textbf{Exceptional prompt.} In the unconditional setting $\epsilon_\theta(\bfx_t, t)$, solving the diffusion ODE (Eq.~(\ref{eqn:flow})) from 0 to $T$ enables us to obtain the better latent code $\bfx_T$ for the real image $\bfx_0$. However, in the text-driven setting $\epsilon_\theta(\bfx_t, t, \mathcal{E})$, existing image editing works \cite{hertz2022prompt, mokady2022null, tumanyan2022plug, wallace2022edict} have shown that the inversion process is prone to significant reconstruction errors, due to the instability induced by CFG \cite{dhariwal2021diffusion, ho2022classifier}:
\begin{align}
	\hat{\epsilon}_\theta(\bfx_t, t, \mathcal{E}, \varnothing) \!=\! s \cdot \epsilon_\theta(\bfx_t, t, \mathcal{E})  \!+\! (1 \!- \! s) \cdot \epsilon_\theta(\bfx_t, t, \varnothing),
	\label{eqn:cfg}
\end{align}
where $\varnothing = \psi("")$ and $\mathcal{E} = \psi(\mathcal{P})$ are embeddings of the null and normal prompt, and $s$ is the guidance scale. Our experiments further reveal that even without CFG, both conditional output $\epsilon_\theta(\bfx_t, t, \mathcal{E})$ and unconditional output $\epsilon_\theta(\bfx_t, t, \varnothing)$ of text-to-image diffusion models still produce large reconstruction errors, as depicted in Figure~\ref{fig:inversion}. 

\begin{figure}[tbp]
	\centering
	\vspace{-0.2cm}
	\includegraphics[width=1\linewidth]{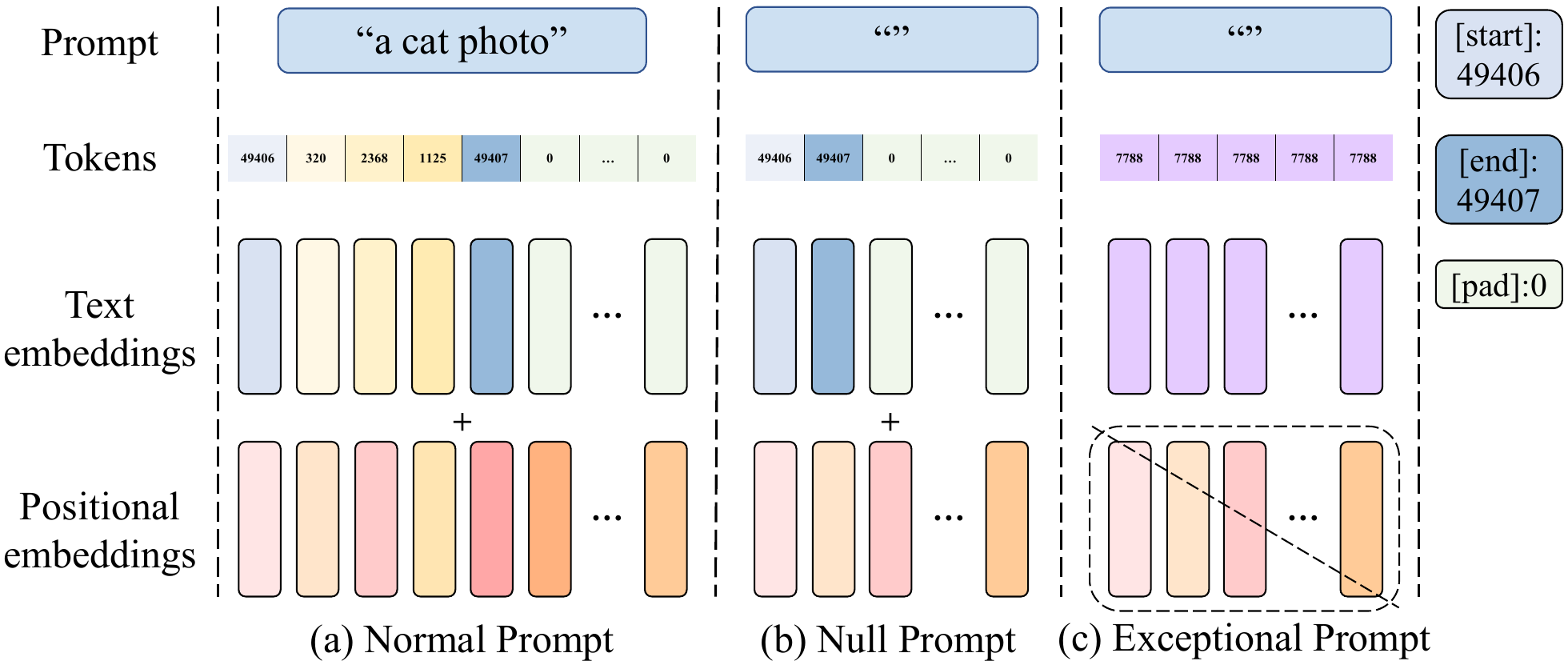}
	\vspace{-0.5cm}
	\caption{The illustration of comparison among (a) Normal Prompt, (b) Null Prompt, and (c) Exceptional Prompt.}
	\vspace{-0.2cm}
	\label{fig:exceptional}
\end{figure}

To achieve accurate inversion, we present a straightforward yet effective solution, namely \textit{exceptional prompt}, $\mathcal{P}_\text{exceptional}$. Intuitively, any information contained within the input prompt can result in the deviation of the backward ODE trajectories from the forward trajectories. Hence, we remove all information by setting all token numbers to a common value and eliminating positional embeddings for the text prompt, as depicted in Figure~\ref{fig:exceptional}. Importantly, the exceptional prompt is distinguished from the null prompt by its absence of special tokens, such as {\tt [startoftext]}, {\tt [endoftext]}, and {\tt [pad]}, which still retain information. The exceptional prompt is applied only in image inversion but not in the composition process. The choice of the token value does not significantly affect the inversion. The detailed analysis of the exceptional prompt and token value selection is provided in Appendix \ref{ep_analysis} and \ref{sec:token_value_analysis}, respectively.
 

Figure~\ref{fig:inversion} visually demonstrates the effectiveness of the exceptional prompt. Our results $\epsilon_\theta(\bfx_t, t, \mathcal{W})$ with the exceptional prompt embedding $\mathcal{W} = \psi(\mathcal{P}_\text{exceptional})$ are more visually accurate than others. The quantitative experiments are shown in Section~\ref{sec:recon}. All the results in Figure~\ref{fig:inversion} are obtained by solving the forward and backward diffusion ODEs using the second-order DPM-Solver++ \cite{lu2022dpmb} in 20 steps. 

\subsection{Training-Free Image Composition}
\label{sec:atten-comp}
Upon equipping the accurate invertibility, image composition can be performed based on it. The composition process consists of two key components: \textbf{noise incorporation} and \textbf{composite self-attention maps injection}. \\

\noindent \textbf{Noise incorporation.} Before inverting images into noises, a simple preprocessing step is necessary for the reference image. Typically, only the foreground in the reference is desired for composition, so the preprocessing step involves using a pretrained segmentation model \cite{zhang2021k} to remove the background, resizing and repositioning the object to match the user's mask in the main image, and padding it with zeros to ensure it is the same size as the main image (See Appendix \ref{sec:preprocessing} for visual illustration). 

Once the preprocessing is complete, the main and reference images are inverted to corresponding noises $\bfx^\text{m}_T$ and $\bfx^\text{r}_T$ by solving diffusion ODEs (Eq.~(\ref{eqn:flow})) from 0 to $T$ with the exceptional prompt $\mathcal{P}_\text{exceptional}$. $\bfx^\text{m}_T$ and $\bfx^\text{r}_T$ are then merged with standard Gaussian noise $\bfz$ to create the starting point $\bfx^*_T$ for generating the composition. Formally, the incorporated noise $\bfx^*_T$ is calculated by
\begin{equation}
	\bfx^*_T = \bfx^\text{r}_T \odot \bfM^\text{seg} + \bfx^\text{m}_T \odot (\bf{1}-\bfM^\text{user}) + \bfz \odot (\bfM^\text{user} \oplus \bfM^\text{seg}),
	\label{eqn:noise_incorportaion}
\end{equation}
where $\bfz \sim \mathcal{N}(\bf{0}, \bf{I})$, $\bfM^\text{user}$ is the user mask, $\bfM^\text{seg}$ is the segmentation mask for reference image, and $\bfM^\text{user} \oplus \bfM^\text{seg}$ is the XOR of them, which is the transition area. The incorporation of $\bfz$ enhances the smoothness of the transition between the regions inside and outside the user mask, by effectively leveraging the prior knowledge of the text-driven diffusion model to inpaint the transition area. Empirically, for cross-domain composition, incorporating the starting point in the noise space usually is more effective, while solely for photorealism, composing in the pixel/latent space and then inverting it as the starting point is more favorable.\\

\noindent \textbf{Composite self-attention map injection.} The incorporated noise $\bfx^*_T$ is employed as the starting point for solving the diffusion ODE from $T$ to 0 with a normal prompt $\mathcal{P}$ to ultimately generate the composition. $\mathcal{P}$ is intended to assist in inpainting transition areas. However, relying solely on noise incorporation, the pretrained text-to-image model cannot preserve the appearance of the main and reference images effectively, as shown in Figure~\ref{fig:ablation1}. To tackle this problem, we propose injecting composite self-attention maps in a specially designed manner, as the semantic information is basically retained within the rows and columns of self-attention maps (See Appendix \ref{sec:sa_visual} for visual illustrations).

\begin{figure}[tbp]
	\centering
	\vspace{-0.5cm}
	\includegraphics[width=0.7\linewidth]{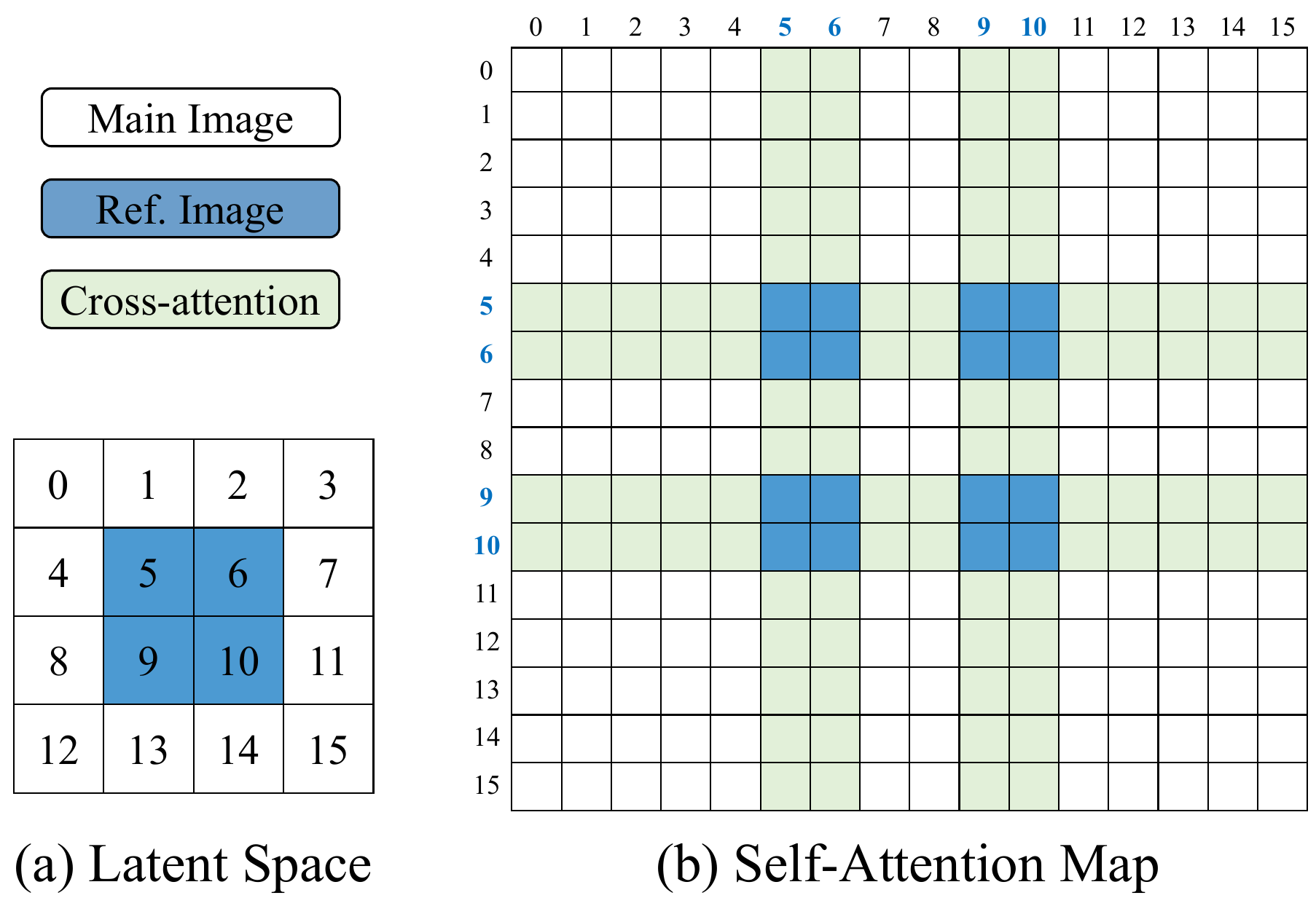}
	\caption{A toy example for attention composition.}
	\vspace{-0.4cm}
	\label{fig:atten_inject}
\end{figure}

The composite self-attention map comprises three constituents: two self-attention maps, $\bfA^\text{m}_{l,t}$ and $\bfA^\text{r}_{l,t}$, corresponding to the main and reference images, and a cross-attention map, $\bfA^\text{cross}_{l,t}$, calculated between them. The composition way is illustrated in Figure~\ref{fig:atten_inject}. To compose the reference image in the blue regions of Figure~\ref{fig:atten_inject}~(a), its self-attention map $\bfA^\text{r}_{l,t}$ should be placed in the corresponding blue regions of Figure~\ref{fig:atten_inject}~(b), since the 5th patch can only attend to the patches 6, 9, and 10 in the context of self-attention. The green regions in Figure~\ref{fig:atten_inject}~(b) should contain the cross-attention map, $\bfA^\text{cross}_{l,t}$, which infuses contextual information from the white regions into the blue regions. If the green regions are preserved as the self-attention of the white regions without replacement, the information stored there only reflects the relation between the original patches, such as the 0th and original 5th patches in the example of index (5,0) or (0,5). This results in a lack of surrounding information being provided to the new 5th patch, such as painting or sketching, which is necessary for seamless object transition to other domains (See ablation in Figure~\ref{fig:ablation1}).

Essential constituents $\bfA^\text{m}_{l,t}, \bfA^\text{r}_{l,t}, \bfA^\text{cross}_{l,t}$ are calculated using self-attention modules of the pretrained Stable Diffusion. Typically, a self-attention module at layer $l$ contains three projection matrices $\bfW^q_l$, $\bfW^k_l, \bfW^v_l$ in the same dimension $\mathbb{R}^{d \times d}$. Denote the spatial features of the main and reference image at timestep $t$ and layer $l$ as $\bff_{l,t}^\text{m} \in \mathbb{R}^{(h \times w) \times d}$ and $\bff_{l,t}^\text{r} \in \mathbb{R}^{(h^\prime \times w^\prime) \times d}$, respectively, where $h^\prime \times w^\prime$ is the size of the reference image after resizing to match the size of the user mask. The queries, keys, and values for each self-attention module are obtained as:
\begin{alignat}{9}
	\label{eqn:qkv}
	& \bfq^\text{m}_{l,t} = && \bff^\text{m}_{l,t} \bfW^q_{l}, \quad && \bfk^\text{m}_{l,t} = && \bff^\text{m}_{l,t} \bfW^k_{l}, \quad && \bfv^\text{m}_{l,t} = && \bff^\text{m}_{l,t} \bfW^v_{l}, \\
	& \bfq^\text{r}_{l,t} = && \bff^\text{r}_{l,t} \bfW^q_{l}, && \bfk^\text{r}_{l,t} = && \bff^\text{r}_{l,t} \bfW^k_{l}, && \bfv^\text{r}_{l,t} = && \bff^\text{r}_{l,t} \bfW^v_{l},
\end{alignat}
where $\bfq^\text{m}_{l,t}$, $\bfk^\text{m}_{l,t}$, $\bfv^\text{m}_{l,t} \in \mathbb{R}^{(h \times w) \times d}$, and $\bfq^\text{r}_{l,t}$, $\bfk^\text{r}_{l,t}$, $\bfv^\text{r}_{l,t} \in \mathbb{R}^{(h^\prime \times w^\prime) \times d}$. Thus, $\bfA^\text{m}_{l,t}, \bfA^\text{r}_{l,t}$, and $\bfA^\text{cross}_{l,t}$ are then calculated and composed as $\bfA^*_{l,t}$ for injection:
\begin{align}
	&\bfA^\text{m}_{l,t} = \text{Softmax}\left(\bfq^\text{m}_{l,t} \cdot (\bfk^\text{m}_{l,t})\tran / \sqrt{d} \right), \\
	&\bfA^\text{r}_{l,t} = \text{Softmax}\left(\bfq^\text{r}_{l,t} \cdot (\bfk^\text{r}_{l,t})\tran / \sqrt{d} \right), \\
	&\bfA^\text{cross}_{l,t} = \text{Softmax} \left(\bfq^\text{m}_{l,t} \cdot (\bfk^\text{r}_{l,t})\tran / \sqrt{d} \right), \\
	&\bfA^*_{l,t} = \vartheta_\text{compose}(\bfA^\text{m}_{l,t}, \bfA^\text{r}_{l,t}, \bfA^\text{cross}_{l,t}),
\end{align}
where $\bfA^\text{m}_{l,t}$ $\in {\mathbb{R}^{(h \times w) \times (h \times w)}}$, $\bfA^\text{r}_{l,t}$ $\in {\mathbb{R}^{(h^\prime \times w^\prime) \times (h^\prime \times w^\prime)}}$, $\bfA^\text{cross}_{l,t}$ $\in {\mathbb{R}^{(h \times w) \times (h^\prime \times w^\prime)}}$, and $\vartheta_\text{compose}$ is the function to bulid composite self-attention maps $\bfA^*_{l,t}$ based on patch indices (Figure~\ref{fig:atten_inject}). 


As a result, three diffusion ODEs are solved simultaneously from $T$ to 0. As depicted in Figure~\ref{fig:overview}~(b), ODEs start from the accurate inverted noises $\bfx^\text{r}_T,\bfx^\text{m}_T$, and the interpolated noise $\bfx^*_T$, respectively. The first two ODEs are solved using the exceptional prompt $\mathcal{P}_\text{exceptional}$ to progressively reconstruct the main and reference, thus allowing for the precise retention of $\bfA^\text{m}_{l,t}$, $\bfA^\text{r}_{l,t}$, and $\bfA^\text{cross}_{l,t}$ at each time step $t$. These attention maps are then composed and injected into the third ODE for generating a natural and cohesive composition with a normal prompt $\mathcal{P}$.

To balance the generation of high-level context and finer details \cite{choi2022perception, kwon2022diffusionb}, we set a threshold $\tau_A$ to determine the time steps for injecting composite self-attention maps in the early stage ($t \in [T \times \tau_A, T]$) and allow the model to explore ODE trajectories through a normal prompt $\mathcal{P}$ in the later stage ($t \in [0, T \times \tau_A]$), guided by the prior of the pretrained model. However, this freedom, without the imposition of attention injection constraints, often results in deviations from the desired background (see Figure~\ref{fig:ablation1}). Thus, similar to \cite{avrahami2022blended}, we set an additional threshold, denoted as $\tau_B$, which regulates the trajectory rectification process. This process entails replacing the regions outside the user mask with the reconstructed main image at various time steps, \ie, $\hat{\bfx}^*_t = \bfx^*_t \odot \bfM^\text{user} + \bfx^\text{m}_t \odot (\bf{1}-\bfM^\text{user})$, where $t \in [T \times \tau_B, T]$. Note that only preserving the background at the final step can lead to noticeable artifacts, as shown in Appendix \ref{sec:background_preserve}. 


\begin{figure*}[htbp]
	\centering
	\includegraphics[width=1\linewidth]{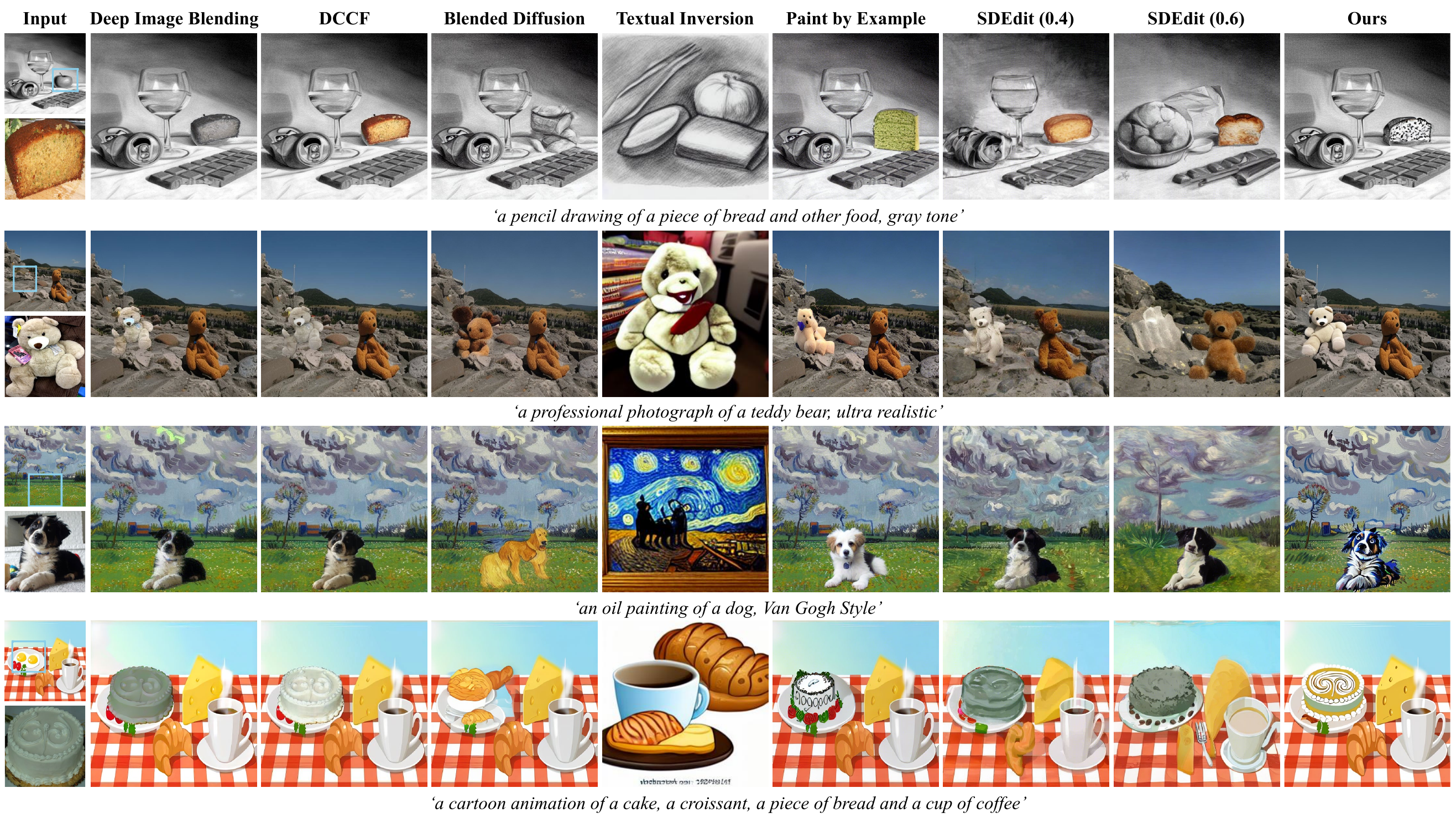}
	\vspace{-0.4cm}
	\caption{Qualitative comparison with SOTA and concurrent baselines in image-guided composition for sketching, photorealism, painting, and cartoon animation domains. Additional results are available in Appendix \ref{sec:addi_qual_results}.}
	\vspace{-0.2cm}
	\label{fig:comparison}
\end{figure*}

\section{Experiments}
This section consists of two sets of experiments. The first set assesses the effectiveness of the exceptional prompt (Section~\ref{sec:recon}). The second set evaluates our image composition framework qualitatively and quantitatively (Section~\ref{sec:comparison}), followed by an ablation study (Section~\ref{sec:ablation}).


\begin{table}[tbp]
	\caption{The reconstruction comparison on CelebA-HQ.}
	\vspace{-0.6cm}
	\small
	\begin{center}
		\resizebox{0.48\textwidth}{!}{
			\begin{tabular}{llccc}
				\toprule
				& Method & MAE $\downarrow$ & LPIPS $\downarrow$ & SSIM $\uparrow$ \\ 
				\midrule
				\multirow{2}{*}{Optimization} & I2S \cite{abdal2019image2stylegan} & 0.064 & 0.134 & 0.872  \\
				& PTI \cite{roich2022pivotal} & 0.062 & 0.132 & 0.877  \\
				\midrule
				\multirow{5}{*}{Encoder} & pSp \cite{richardson2021encoding} & 0.079 & 0.169 & 0.793  \\
				& e4e \cite{tov2021designing} & 0.092 & 0.221 & 0.742  \\
				& ReStyle w/ pSp \cite{alaluf2021restyle} & 0.073 & 0.145 & 0.823  \\
				& ReStyle w/ e4e \cite{alaluf2021restyle} & 0.089 & 0.202 & 0.758  \\
				& HFGI w/ e4e \cite{wang2022high} & 0.062 & 0.127 & 0.877  \\
				\midrule
				\multirow{5}{*}{Diffusion} & SD w/ CFG & 0.134 & 0.340 & 0.637  \\
				& SD w/ Cond. & 0.126 & 0.308 & 0.654 \\
				& SD w/ Uncond. & 0.126 & 0.304 & 0.655 \\
				& DiffusionCLIP \cite{kim2022diffusionclip} & 0.020 & 0.073 & 0.914 \\
				& Ours & \textbf{0.019} & \textbf{0.047} & \textbf{0.918}  \\
				\midrule
				Upper Bound & VQAE \cite{esser2021taming} & 0.018 & 0.043 & 0.919 \\
				\bottomrule
			\end{tabular}
		}
	\end{center}
	\label{tab:recon_celeb}
\end{table}

\begin{table}[tbp]
	\vspace{-0.4cm} 
	\caption{The further reconstruction comparison on COCO and ImageNet. *: an upper bound.}
	\vspace{-0.6cm}
	\begin{center}
		\resizebox{0.48\textwidth}{!}{
			\begin{tabular}{lcccccc}
				\toprule
				\multirow{2}{*}{Method} & \multicolumn{3}{c}{MSCOCO (5000)} & \multicolumn{3}{c}{ImageNet (3000)} \\
				\cmidrule(r){2-4}  \cmidrule(r){5-7} 
				& MAE $\downarrow$ & LPIPS $\downarrow$ & SSIM $\uparrow$ & MAE $\downarrow$ & LPIPS $\downarrow$ & SSIM $\uparrow$\\
				\midrule
				SD w/ CFG & 0.150 & 0.458 & 0.568 & 0.132 & 0.496 & 0.575 \\
				SD w/ Cond. & 0.122 & 0.359 & 0.633 & 0.109 & 0.389 & 0.645\\
				SD w/ Uncond. & 0.120 & 0.363 & 0.636 & 0.114 & 0.406 & 0.635\\
				Ours & \textbf{0.030} & \textbf{0.073} & \textbf{0.868} & \textbf{0.033} & \textbf{0.087} & \textbf{0.852}\\
				\midrule
				VQAE* \cite{esser2021taming} & 0.030 & 0.069 & 0.870 & 0.032 & 0.084 & 0.854 \\
				\bottomrule
		\end{tabular}}
	\end{center}
	\vspace{-0.7cm} 
	\label{tab:recon}
\end{table}

\subsection{Image Reconstruction} 
\label{sec:recon}
To assess the effectiveness of the exceptional prompt, we compared its performance with SOTA GAN \cite{abdal2019image2stylegan, alaluf2021restyle, richardson2021encoding, roich2022pivotal, tov2021designing, wang2022high} and diffusion \cite{kim2022diffusionclip} inversion methods on the CelebA-HQ \cite{karras2017progressive}, following the same setting as described in \cite{kim2022diffusionclip, wang2022high}. Additionally, we conducted experiments on the ImageNet \cite{deng2009imagenet} and COCO \cite{lin2014microsoft} with Stable Diffusion to further validate our findings. Our results (Tables~\ref{tab:recon_celeb} and \ref{tab:recon}) show that the exceptional prompt is highly effective in producing reconstructions that closely approximate the upper bound established by the vector quantized autoencoder (VQAE) \cite{esser2021taming} across all metrics, including MAE, LPIPS \cite{zhang2018perceptual}, and SSIM. The qualitative comparison is shown in Figure~\ref{fig:inversion}. All results of Stable Diffusion, including ours, are sampled by the second-order DPM-Solver++ \cite{lu2022dpmb}. Experimental settings are detailed in Appendix \ref{sec:exp_settings}.

\subsection{Image Composition Comparisons}
\label{sec:comparison}

\noindent \textbf{Test benchmark.} As there is currently no benchmark for testing cross-domain image-guided composition as a whole, we developed a test benchmark containing 332 samples. Each sample in the benchmark consists of a main (background) image, a reference (foreground) image, a user mask, and a text prompt. The main images comprise four visual domains: photorealism, pencil sketching, oil painting, and cartoon animation. All reference images are from the photorealism domain as the reference requires segmentation models, which are generally more effective in this domain. Further details are available in Appendix \ref{benchmark}.\\


\begin{table}[tbp]
	\caption{Quantitative evaluation results for image composition in the photorealism domain.}
	\vspace{-0.6cm}
	\begin{center}
		\resizebox{0.48\textwidth}{!}{
			\begin{tabular}{lcccc}
				\toprule
				Method & $\text{LPIPS}_\text{(BG)}\downarrow$ & $\text{LPIPS}_\text{(FG)}\downarrow$ & $\text{CLIP}_\text{(Image)}\uparrow$ & $\text{CLIP}_\text{(Text)}\uparrow$ \\
				\midrule
				SDEdit (0.4) \cite{meng2021sdedit} & 0.35 & 0.62 & 80.56 & 27.73 \\
				SDEdit (0.6) \cite{meng2021sdedit} & 0.42 & 0.66 & 77.68 & 27.98 \\
				Blended \cite{avrahami2022blended_latent} & 0.11 & 0.77 & 73.25 & 25.19 \\
				Paint \cite{yang2022paint} & 0.13 & 0.73 & 80.26 & 25.92 \\ 
				DIB \cite{zhang2020deep} & 0.11 & 0.63 & 77.57 & 26.84 \\
				Ours & \textbf{0.10} & \textbf{0.60} & \textbf{82.86} & \textbf{28.11}\\
				\bottomrule
		\end{tabular}}
	\end{center}
	\vspace{-0.7cm} 
	\label{tab:quan}
\end{table}

\noindent \textbf{Qualitative comparisons.} 
Our qualitative comparisons are performed across four visual domains, employing SOTA and concurrent baselines that are applicable to image-guided composition, including Deep Image Blending (DIB) \cite{zhang2020deep}, DCCF \cite{xue2022dccf}, Blended Diffusion \cite{avrahami2022blended_latent}, Textual Inversion \cite{gal2022image}, Paint by Example \cite{yang2022paint}, and SDEdit \cite{meng2021sdedit} under two different noising levels. As shown in Figure~\ref{fig:comparison}, our framework is capable of seamlessly composing objects into various domains while maintaining their identities. In contrast, DIB and DCCF fall short in processing the transition areas, leading to noticeable artifacts. Blended Diffusion's foreground generation and Textual Inversion's background generation rely solely on text prompts, causing deviations from the user's intention. While Paint by Example effectively composes images within its photorealistic training domain, it struggles to adapt to other domains. Additionally, SDEdit with fewer timesteps is suitable for image composition in terms of preserving the identifying features of the reference, but the background is changed. See Appendix \ref{sec:addi_qual_results} for additional comparisons.\\

\noindent \textbf{Quantitative analysis.}
The baselines are primarily trained in the photorealism domain, where the objective metrics are more effective; therefore, we focused our quantitative comparison within this domain and relied on user study for comparison in other domains. We assess the same baselines as in the qualitative comparison, with the exception of Textual Inversion \cite{gal2022image}, which involves instance-based optimization, and DCCF \cite{xue2022dccf}, which is used for harmonizing images after copy-and-paste operations. Four metrics are considered: (1) $\text{LPIPS}_\text{(BG)}$ \cite{zhang2018perceptual} measures the background consistency, (2) $\text{LPIPS}_\text{(FG)}$ \cite{zhang2018perceptual} evaluates the low-level similarity between the edited region and the reference foreground, (3) $\text{CLIP}_\text{(Image)}$ \cite{radford2021learning} evaluates the semantic similarity between the edited region and the reference in the CLIP embedding space, and (4) $\text{CLIP}_\text{(Text)}$ \cite{radford2021learning} measures the semantic alignment between the text prompt and the resultant image. As presented in Table~\ref{tab:quan}, our method outperforms all baselines. We achieve well preservation of the background, high object correspondence in both low-level and high-level feature spaces, as well as a high degree of alignment with the text prompt. \\

\noindent \textbf{User study.} 
We conducted a user study to compare image composition baselines across domains. We recruited 50 participants via Amazon and tasked them with completing 40 ranking questions. Each question consists of 5 options, generated using distinct methods. Details are available in Appendix \ref{user}. The ranking criteria comprehensively considered foreground preservation, background consistency, seamless composition, and text alignment. The results are listed in Table~\ref{tab:user_study}, where the domain information is presented in the format of `foreground domain \& background domain', \eg, photorealism \& oil painting. Our method was favored by most participants across different domains.


\begin{table}[tbp]
	\scriptsize
	\caption{User study: higher score, better ranking. P: photorealism; O: oil painting; S: sketchy painting; C: cartoon.}
	\vspace{-0.4cm}
	\begin{center}
			\resizebox{0.48\textwidth}{!}{
			\begin{tabular}{lccccc}
					\toprule
					Method & P \& P & P \& O & P \& S & P \& C & Total \\
					\midrule
					Blended \cite{avrahami2022blended_latent} & 1.807 & 2.314 & 2.680 & 2.100 & 2.093 \\
					SDEdit (0.6) \cite{meng2021sdedit} & 2.368 & 3.063 & 2.409 & 2.713& 2.485 \\
					Paint \cite{yang2022paint} & 2.879 & 2.306 & 2.043 & 2.673 & 2.666 \\
					DCCF \cite{xue2022dccf} & 3.838 & 3.237 & 3.297 & 3.470 & 3.583\\
					Ours & \textbf{4.108} & \textbf{4.080} & \textbf{4.571} & \textbf{4.043} & \textbf{4.175} \\
					\bottomrule
			\end{tabular}}
		\end{center}
	\vspace{-0.2cm} 
	\label{tab:user_study}
\end{table}

\begin{table}[tbp]
	\caption{Ablation study: quantitative comparison of various variants of our framework.}
	\vspace{-0.4cm}
	\begin{center}
		\resizebox{0.48\textwidth}{!}{
			\begin{tabular}{lcccc}
				\toprule
				Config & $\text{LPIPS}_\text{(BG)}\downarrow$ & $\text{LPIPS}_\text{(FG)}\downarrow$ & $\text{CLIP}_\text{(Image)}\uparrow$ & $\text{CLIP}_\text{(Text)}\uparrow$ \\
				\midrule
				Baseline & 0.34 & 0.65 & 75.13 & \textbf{29.00} \\
				+ $\mathcal{P}_\text{exceptional}$ & 0.32 & 0.64 & 78.54 & 28.23\\
				+ SA injection & 0.25 & 0.61 & 81.64 & 28.58 \\
				+ CA injection & 0.26 & 0.60 & 81.63 & 28.52 \\ 
				+ Background & \textbf{0.10} &  \textbf{0.60} & \textbf{82.86} & 28.11 \\
				\bottomrule
		\end{tabular}}
	\end{center}
	\vspace{-0.4cm} 
	\label{tab:ablation}
\end{table}

\subsection{Ablation Study}
\label{sec:ablation}
We ablate our key design choices in the following cases: (1) Baseline, where the composition is generated by solving the diffusion ODE from $T$ to 0 using DPM-Solver++ without any injection. The starting point is composed by inverted noises under the normal prompt; (2) The exceptional prompt is applied to obtain accurate inverted noises; (3) The self-attention maps $\bfA^\text{m}_{l,t}$ and $\bfA^\text{r}_{l,t}$ are composed and then injected; (4) The cross-attention $\bfA^\text{cross}_{l,t}$ between the main and reference images is further composed for injection; (5) The background preservation is applied.

\begin{figure}[tbp]
	\centering
	\includegraphics[width=1\linewidth]{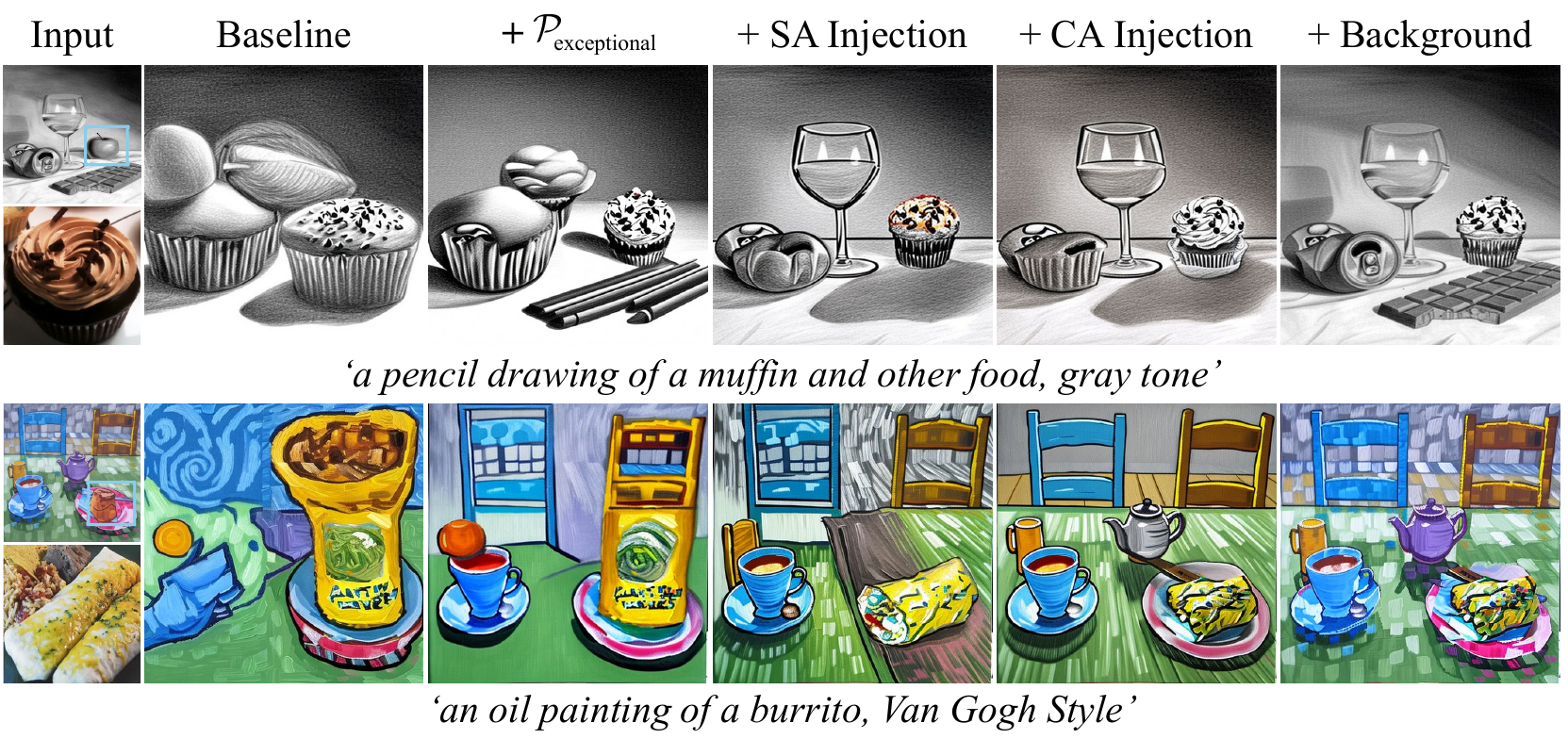}
	\vspace{-0.4cm}
	\caption{Ablation study of different variants of our framework. SA: self-attention. CA: cross-attention.}
	\vspace{-0.5cm}
	\label{fig:ablation1}
\end{figure}

Table~\ref{tab:ablation} presents the quantitative results, which indicate that the complete algorithm outperforms other variants in all metrics except for $\text{CLIP}_\text{(Text)}$. Notably, the baseline achieves the best $\text{CLIP}_\text{(Text)}$ as it generates compositions solely relying on the normal prompt without any extra constraint. While metrics alone may not reveal the complete effectiveness of cross-attention injection, Figure~\ref{fig:ablation1} illustrates that the interactions between the main and reference images are highly beneficial to both foreground and background in terms of preserving appearance and switching domains. Note that preserving the background at different noise levels affects both background and foreground, which is distinct from preserving solely at the final step (Appendix \ref{sec:background_preserve}). Additional ablation results are shown in Appendix \ref{sec:addi_qual_results}.

\section{Limitations and Future Work}
The primary limitation of our work lies in its inability to generate an object view that critically differs from the given reference. As a result, the choice of reference image may be restricted at times. This is because the model relies on self-attention maps to provide layout and appearance information, which in turn constrains the development of alternative views. While introducing a loose self-attention injection can generate different views, this often compromises the preservation of the object's appearance. To overcome this, further research could explore utilizing personalized concept learning techniques, such as Textual Inversion \cite{gal2022image}, to encode identity information in the text prompt through special embeddings. Alternatively, utilizing NeRF-relevant techniques \cite{cai2022diffdreamer, mildenhall2021nerf, wang2022score} can generate other views of the object for a specific scene, but this can require expensive training. Furthermore, due to the fact that our approach relies on Stable Diffusion, it inherits its shortcomings and biases, which may result in producing artifacts in certain scenarios.

\section{Conclusion}
We introduced a method that leverages the high-order ODE solver with the exceptional prompt to achieve precise inversion of real images, which serves as a foundation for further manipulation. Building upon this, we propose a novel training-free framework, TF-ICON, that enables attention-based text-to-image diffusion models to perform image-guided composition across different domains. Our experimental results demonstrate that our approach outperforms the SOTA baselines for both image inversion and composition. We believe that image composition has the potential to become an essential tool for content creators, offering significant benefits for downstream applications in various industries.

\section*{Acknowledgement}
This research is supported by the National Research Foundation, Singapore under its Strategic Capability Research Centres Funding Initiative. Any opinions, findings and conclusions or recommendations expressed in this material are those of the author(s) and do not reflect the views of National Research Foundation, Singapore.

{\small
	\bibliographystyle{ieee_fullname}
	\bibliography{egbib}
}

\clearpage
\appendix

\noindent\textbf{\Large Appendix}
\renewcommand\thesection{\Alph{section}}
\setcounter{section}{0}

\section{Additional Analysis}
\label{sec:Additional Analysis}
\subsection{Better Latent Code with ODE Solvers}
\label{sec:better_latent_DPM}
In Section~\ref{sec:exceptional prompt}, we argue that the utilization of diffusion ODE solvers \cite{karras2022elucidating, liu2022pseudo, lu2022dpma, lu2022dpmb} as encoders, commencing from the real image $\bfx_0$, results in better latent representation $\bfx_T$ as compared to those acquired via the commonly used DDIM \cite{song2020denoising}. This improvement is attributed to the better alignment between the forward and backward ODE trajectories produced by higher-order ODE solvers. 

\begin{figure}[tbp]
	\centering
	\includegraphics[width=1\linewidth]{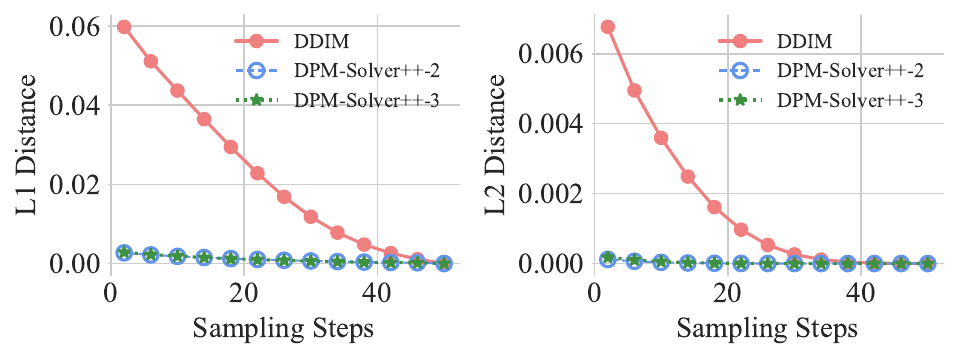}
	\caption{The comparison of the alignment of forward and backward trajectories from DDIM inversion and high-order DPM-Solver++. L1 and L2 distances were computed at each time step between the forward and backward intermediates, and then averaged over 5000 images. The curves representing the second and third order DPM-Solver++ are almost overlapping. Please zoom in for a closer look.}
	\label{fig:DPM}
\end{figure}

This claim is supported by the experimental results presented in Figure~\ref{fig:DPM}. Specifically, we performed image inversion on the COCO2017 \cite{lin2014microsoft} validation set of 5000 images using DDIM, as well as the second and third order DPM-Solver++ \cite{lu2022dpmb}. This involved encoding real images into noises and subsequently decoding the noises back into real images, with both procedures consisting of 50 steps. The forward and backward intermediate states were preserved as $\{\bfx_0^\text{enc},\bfx_1^\text{enc},\cdots,\bfx_{50}^\text{enc}\}$ and $\{\bfx_0^\text{dec},\bfx_1^\text{dec},\cdots,\bfx_{50}^\text{dec}\}$, respectively. L1 and L2 distances between the forward and backward processes' intermediates were computed at each time step. Figure~\ref{fig:DPM} presents the average values of the distances. 

The experimental results demonstrate that the higher-order DPM-Solver++ exhibits a smaller difference between the forward and backward intermediates, signifying better alignment between the forward and backward trajectories, in comparison to DDIM, which is equivalent to a first-order solver. Furthermore, the experimental results suggest that an increase in the order of the ODE solver does not lead to additional improvement in alignment. 

\begin{figure}[tbp]
	\centering
	\includegraphics[width=1\linewidth]{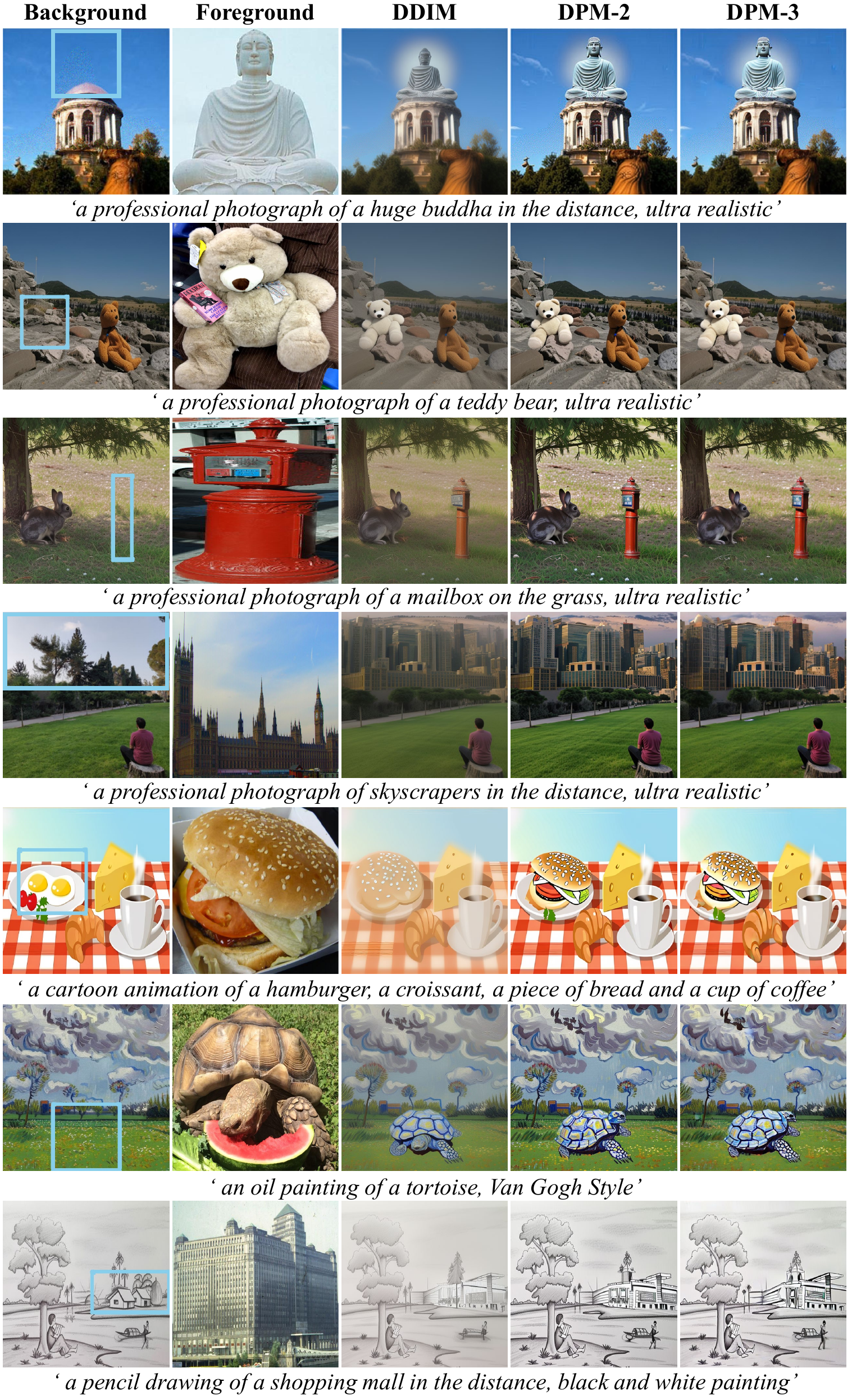}
	\caption{The visual comparison between image compositions achieved through the utilization of high-order DPM solvers++ and DDIM inversion. The image compositions resulting from DDIM inversion exhibit more blurring when compared to those generated by high-order DPM solvers++ employing the same 20-step sampling process. Augmenting the solver's order does not result in noteworthy visual enhancements.}
	\vspace{-0.2cm}
	\label{fig:DPM_vs_DDIM}
\end{figure}

Figure~\ref{fig:DPM_vs_DDIM} presents a visual comparison between image compositions achieved through the utilization of high-order DPM solvers and DDIM inversion. Due to subpar alignment between forward and backward intermediates, The inversion codes of DDIM yield blurred images when using the same sampling steps (20 steps).

Given that Lu \etal \cite{lu2022dpmb} has established the suitability of the second-order DPM-Solver++ for guided sampling\footnote {\scriptsize \url{https://github.com/LuChengTHU/dpm-solver}}, we employed the second-order one for all the experiments conducted with TF-ICON.

\subsection{Exceptional Prompt Analysis}
\label{ep_analysis}
Denote the image features as \(\mathbf{f} \in \mathbb{R}^{s\times d_1}\) and the embedding of the exceptional prompt as \(\mathbf{T} \in \mathbb{R}^{l \times d_2}\), where \(d_1, d_2\) denote the dim of the image and text embeddings, \(s=h \times w\) is the res of the latent space, and \(l\) is the maximum length of the prompt. By assigning the same value to all tokens and discarding the positional embeddings, each row of \(\mathbf{T}\) is identical. In a cross-attention module, we have \(\mathbf{W}^q \in \mathbb{R}^{d_1 \times d_1}\), \(\mathbf{W}^k, \mathbf{W}^v \in \mathbb{R}^{d_2 \times d_1}\) and $\mathbf{q} = \mathbf{f} \cdot \mathbf{W}^q, \mathbf{k} = \mathbf{T} \cdot \mathbf{W}^k, \mathbf{v} = \mathbf{T} \cdot \mathbf{W}^v. $ 
When a matrix with identical rows multiplies another matrix, the resultant matrix also exhibits identical rows. Thus, \(\mathbf{k}, \mathbf{v}\) have identical rows, and \(\mathbf{q} \cdot \mathbf{k}\tran\) has identical columns. Applying the softmax row-wise to $\mathbf{q} \cdot \mathbf{k}\tran$ generates a constant attention map \(\mathbf{A}  = \frac{1}{l} \cdot \mathbf{1}_{s\times l} \). The output \(\mathbf{o} = \mathbf{A} \cdot \mathbf{v}\) hence exhibits identical rows and is then added to the input, \ie, $\mathbf{f} + \mathbf{o}$, before moving to the next layer. Each row of $\mathbf{f}_{s \times d_1}$ is the embedding of each patch. In the exceptional prompt, all patch embeddings experience a consistent directional movement, but normal and null prompts with varying row vectors cause embeddings to move in various directions, thereby disrupting the image pattern.

\subsection{Elaboration of Inversion Results}
\label{sec:elab_for_fig3}
Two specific points in Figure~\ref{fig:inversion} require attention. Firstly, it is true that CFG typically amplifies instability, resulting in subpar metrics (Tables~\ref{tab:recon_celeb} and \ref{tab:recon}), while satisfactory reconstruction from CFG output is possible, albeit less common, even with only DDIM (Figure 10 in \cite{hertz2022prompt} and 3rd row of Figure~\ref{fig:inversion}). Secondly, the unconditional output does not necessarily outperform CFG or conditional one, as the unconditional/null prompt contains special symbols (Figure~\ref{fig:exceptional}), which also add information and lead to inconsistent directional shifts in image embeddings (See Section~\ref{ep_analysis}). Thus, the unconditional output may perform poorly than others (4th and 6th row of Figure~\ref{fig:addi_imagenet}). Figure~\ref{fig:inversion} shows unconditional (1st row), conditional (2nd row), or CFG (3rd row) output can yield the best reconstruction among them.

\begin{figure*}[tbp]
	\centering
	\includegraphics[width=1\linewidth]{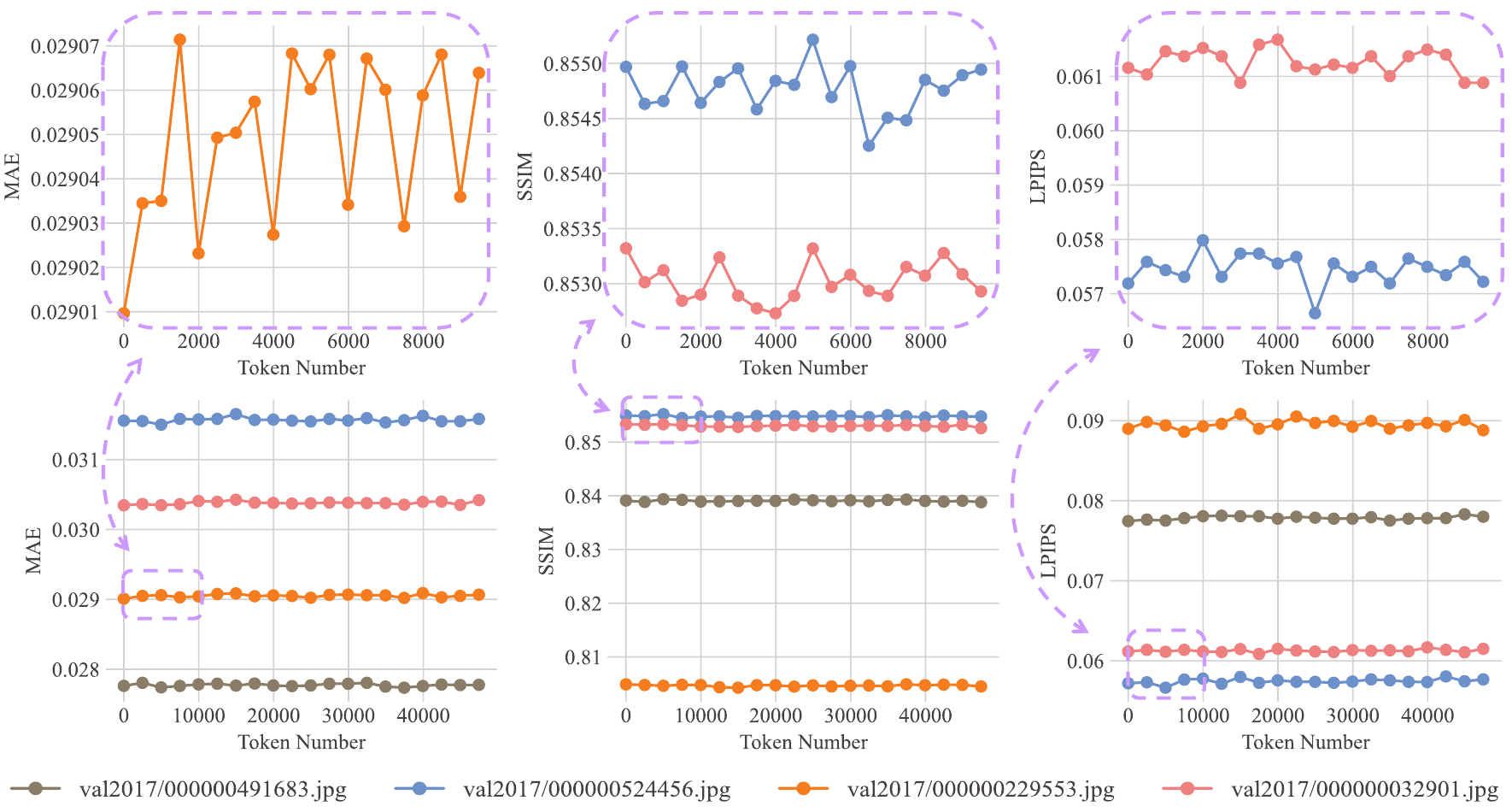}
	\caption{The analysis of the impact of the common token values in the exceptional prompt. The first row displays \textcolor[RGB]{119,74,169}{a magnified view} of an area from the second row. For each image randomly sampled from the COCO, the exceptional prompt is applied with 100 uniformly sampled token values on Stable Diffusion to perform image inversion. The inversion metrics, including MAE, SSIM, and LPIPS, exhibit negligible variations as the token value is modified.}
	\label{fig:tokens}
\end{figure*}

\subsection{Token Value Analysis}
\label{sec:token_value_analysis}
In Section~\ref{sec:exceptional prompt}, we contend that the choice of token value has no significant impact on the inversion performance. To justify this, we uniformly sampled 100 token values from the set of 49407 values and employed them as the common token value in the exceptional prompt $\mathcal{P}_\text{exceptional}$. All experimental results are obtained using Stable Diffusion \cite{rombach2022high} with the exceptional prompt (100 different token values), sampled through the second-order DPM-Solver++ in 50 steps. Three metrics, namely MAE, LPIPS, and SSIM, were used to assess the inversion performance. 

The experimental results for four randomly sampled images from the COCO2017 validation set are shown in Figure~\ref{fig:tokens}. The top row of Figure~\ref{fig:tokens} displays a magnified view of a specific area from the second row. Notably, for a single image, each token value produces nearly identical inversion performance, with only minor fluctuations occurring within a narrow range.

Furthermore, we randomly sampled 150 images from the COCO2017 validation set. For each image, we calculated the means and standard deviations of the three metrics among the reconstruction results of the 100 tokens. The metrics were averaged over 150 images, as listed in Table~\ref{tab:tokens}. Importantly, the average standard deviations of all metrics for the reconstructions of different tokens are remarkably low, indicating that the selection of token values does not significantly affect the performance of inversion.

\begin{table}[tbp]
	\small
	\caption{Means and standard deviations of metrics among the reconstruction results of 100 tokens, averaged over 150 images randomly sampled from the COCO.}
	\vspace{-0.3cm}
	\begin{center}
		\resizebox{0.48\textwidth}{!}{
			\begin{tabular}{llll}
				\toprule
				& MAE & LPIPS & SSIM \\
				\midrule
				Mean & 0.0323 & 0.0703 & 0.8560  \\
				Standard Deviation & 9.22$\times 10^{-5}$  & 9.29$\times 10^{-4}$ & 3.81$\times 10^{-4}$ \\
				\bottomrule
		\end{tabular}}
	\end{center}
	\vspace{-0.4cm}
	\label{tab:tokens}
\end{table}

\section{Implementation Details}
\subsection{Preprocessing}
\label{sec:preprocessing}
\begin{figure}[tbp]
	\centering
	\includegraphics[width=1\linewidth]{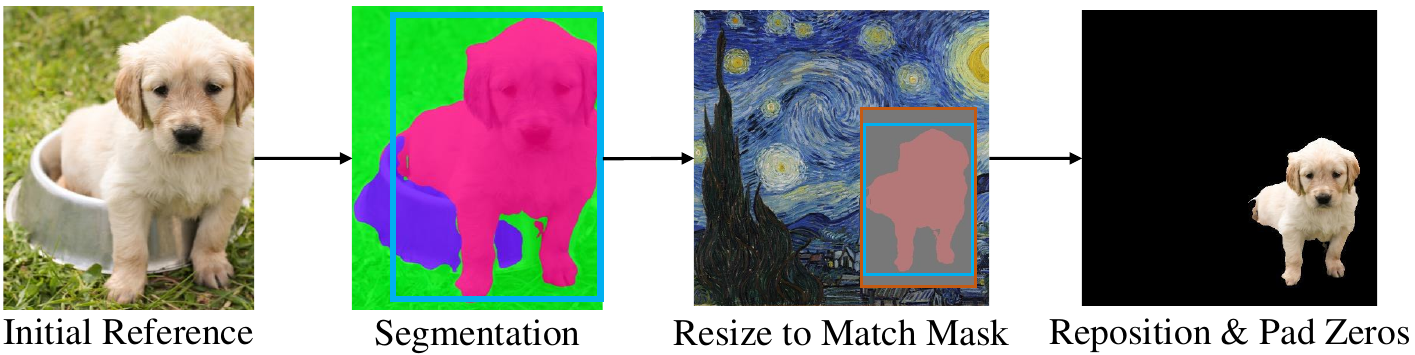}
	\caption{The preprocessing pipeline for the reference image. (1) The centralized reference image is initially processed by a pretrained segmentation model; (2) the segmented object region is then extracted, and its dimension is adjusted to match the size of the user mask; (3) the resized image is finally repositioned and padded with zeroes to match the main image's dimension. }
	\label{fig:preprocessing}
\end{figure}

Figure~\ref{fig:preprocessing} illustrates the preprocessing process. Typically, only the foreground in the reference image is desired for composition, so a pretrained segmentation model \cite{zhang2021k} is utilized to segment the object from the background. Next, the extracted object is resized and repositioned to correspond with the user's mask in the main image. Finally, zero padding is applied to the object to ensure it is the same size as the main image.

\begin{figure}[tbp]
	\centering
	\includegraphics[width=1\linewidth]{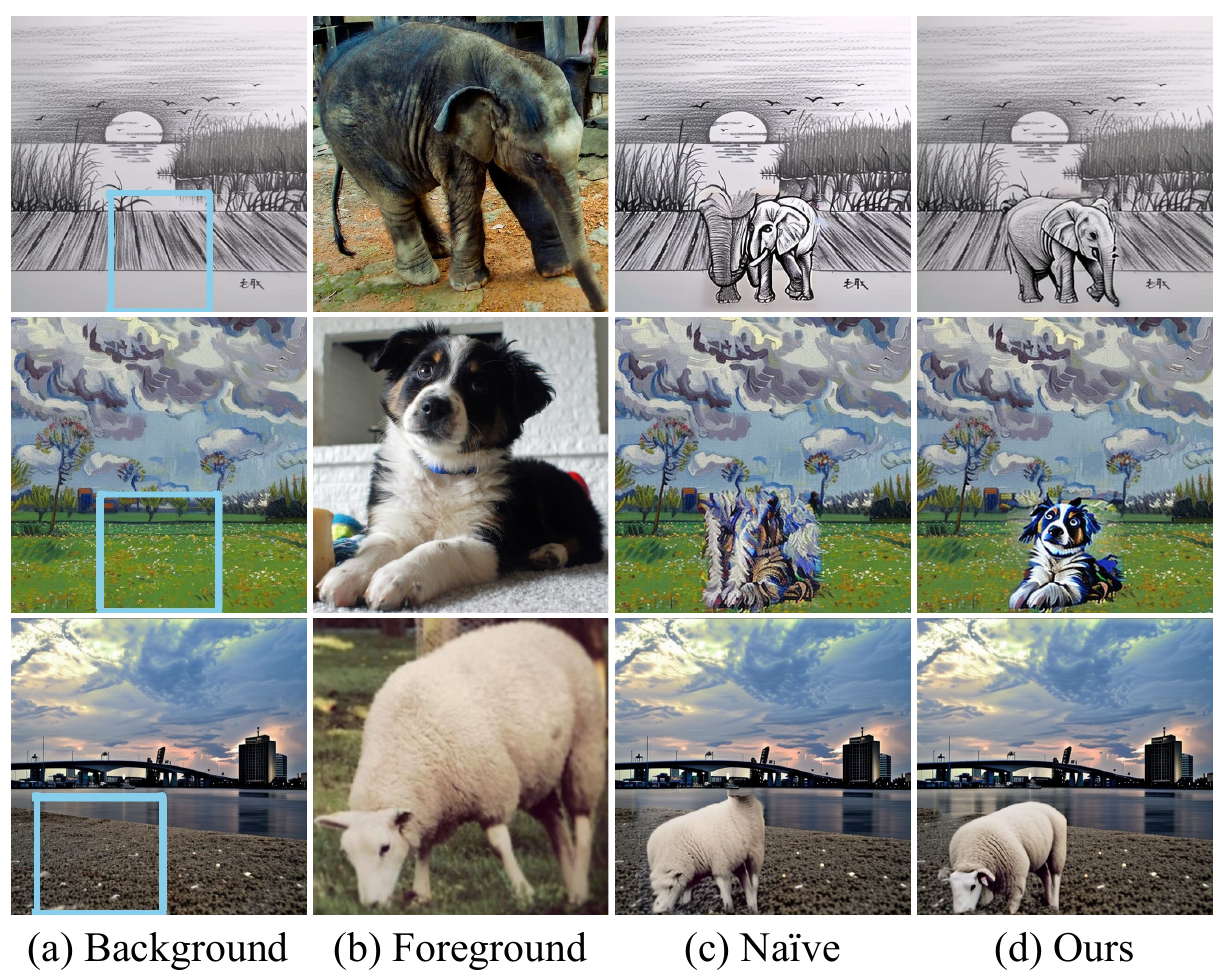}
	\caption{The comparison between two implementations of background preservation. Naïve implementation only preserves the background at the final step, while ours gradually blends background information at various time steps.}
	\label{fig:background}
\end{figure}

\subsection{Algorithm and Running Time}
\label{sec:algo}
Algorithm~\ref{alg:composition} describes the pseudocode of the proposed training-free image composition framework (TF-ICON). The synthesis time for a single image using one A100 GPU card is around 8 seconds, depending on the size of the user mask and reference image.

\captionsetup[algorithm]{font=small}
\begin{algorithm}[tbp]
	\small
	\caption{Training-Free Image Composition}
	\label{alg:composition}
	\begin{algorithmic}[1]
		\State \textbf{Input:} The embeddings of the normal prompt and the exceptional prompt $\mathcal{E} = \psi(\mathcal{P})$ and $\mathcal{W} = \psi(\mathcal{P}_\text{exceptional})$, the main image $\bfI^\text{m}$, the reference image ${{\bfI}^\text{r}}$, the user mask ${\bf{M}}^\text{user}$, the segmentation mask ${\bf{M}}^\text{seg}$, thresholds $\tau_A$, $\tau_B$
		
		\State \textbf{Output:} The composition result ${\bfI}^*$
		\vspace{1mm} \hrule \vspace{1mm}
		
		\State {\tt // Step 1: Starting Point Incorporation }
		
		\State$\bfx^\text{m}_0 = \text{VQ-Encoder} \left(\bfI^\text{m}\right)$; $\bfx^\text{r}_0 = \text{VQ-Encoder} \left(\bfI^\text{r}\right)$
		\For{$t=1, \dotsc, T$}
		\State $\bfx^\text{m}_t \gets $ DPM-Solver++ $\left(\bfx^\text{m}_{t-1}, t-1, \mathcal{W}\right)$
		\State $\bfx^\text{r}_t \gets $ DPM-Solver++ $\left(\bfx^\text{r}_{t-1}, t-1, \mathcal{W}\right)$
		\EndFor
		
		\State $\bfz \sim \mathcal{N}(\bf{0}, \bf{I})$
		\State $\bfx^*_T \gets \bfx^\text{r}_T \odot \bfM^\text{user} + \bfx^\text{m}_T \odot (\bf{1}-\bfM^\text{user}) + \bfz \odot (\bfM^\text{user} \oplus \bfM^\text{seg})$ 
		
		\vspace{1mm}
		\State {\tt // Step 2: Image Composition }
		\For{$t=T, \dotsc, 1$}
		
		
		\State $\bfx^\text{m}_{t-1}, \left\{{\bfA}^\text{m}_t\right\} \gets $ DPM-Solver++ $\left(\bfx^\text{m}_t, t, \mathcal{W}\right)$
		
		
		\State $\bfx^\text{r}_{t-1}, \left\{{\bfA}^\text{r}_t\right\} \gets $ DPM-Solver++ $\left(\bfx^\text{r}_t, t, \mathcal{W}\right)$
		
		\State $\left\{ {\bfA}^\text{cross}_t \right\} \leftarrow \text{CrossAtten}( \bfx^\text{m}_t, \bfx^\text{r}_t)$
		\State $\left\{\bfA^*_{t}\right\} \gets \vartheta_\text{compose}\left(\left\{{\bfA}^\text{m}_t\right\}, \left\{{\bfA}^\text{r}_t\right\}, \left\{{\bfA}^\text{cross}_t\right\}\right)$
		\If{$t > \text{int}(\tau_A \times T)$}
		\State $\bfx^*_{t-1} \gets \text{DPM-Solver++}\left(\bfx^*_t, t, \mathcal{E}, \left\{\bfA^*_{t}\right\}\right)$
		\Else
		\State $\bfx^*_{t-1} \gets \text{DPM-Solver++}\left(\bfx^*_t, t, \mathcal{E}\right)$
		\EndIf
		
		\If{$t > \text{int}(\tau_B \times T)$}
		\State ${\bfx}^*_{t-1} \gets \bfx^*_{t-1} \odot \bfM^\text{user} + \bfx^\text{m}_{t-1} \odot (\bf{1}-\bfM^\text{user})$
		\EndIf
		\EndFor
		\State ${\bfI}^* = \text{VQ-Decoder}(\bfx^*_0)$
		\State \textbf{return} ${\bfI}^*$
	\end{algorithmic}
\end{algorithm}

\subsection{Background Preservation}
\label{sec:background_preserve}
As discussed in Sections~\ref{sec:atten-comp} and \ref{sec:ablation}, preserving the background during denoising should be done gradually at different levels of noise. Preserving the background only at the final time step may result in noticeable artifacts. Figure~\ref{fig:background} provides a comparison between the naïve implementation, which preserves the background only at the final step, and our implementation, which follows a gradual way. The naïve implementation results in obvious artifacts, while ours successfully produces high-quality results. 

The rationale behind this phenomenon is that when two noisy images are blended at a certain noise level, the resulting image may lie outside the targeted manifold. The subsequent steps of diffusion can rectify this issue by moving it toward the next level manifold, thereby gradually improving the coherence of the image. However, if the blending is only performed at the final step in a simplistic manner, the image cannot be corrected any further.

\subsection{Experimental Settings and Hyperparameters}
\label{sec:exp_settings}
\noindent \textbf{Image Reconstruction.} To conduct inversion experiments on the CelebA-HQ \cite{karras2017progressive} (\ie, Table~\ref{tab:recon_celeb}), we followed the experimental settings outlined in \cite{kim2022diffusionclip, wang2022high}. The first 1500 images from the CelebA-HQ were inverted, and the quality of reconstruction from the inverted latent was evaluated using MAE, LPIPS, and SSIM metrics. All Stable Diffusion results were sampled in 50 steps using the second-order DPM-Solver++. The normal prompt for the conditional output and the output with CFG was set as \textit{`a portrait photo'}. The CFG scale was 5. The common token value of the exceptional prompt was 7788.

In further experiments on the COCO2017 (\ie, Table~\ref{tab:recon}), the entire validation set with 5000 images was used. The first listed caption of each image in the annotations serves as the normal prompt. In the experiments on the ImageNet \cite{deng2009imagenet} (\ie, Table~\ref{tab:recon}), 3000 images were randomly sampled from the ImageNet validation set. \textit{`a photo of the [class]'} was used as the normal prompt. For both datasets, the CFG scale was set at 5, and the common token value of 7788 was used in the exceptional prompt. \\

\noindent \textbf{Image Composition.} Since most baselines are trained only in the photorealism domain, where objective metrics are more effective, we conducted our quantitative comparison in this domain. However, for other domains, we relied on user study and qualitative comparisons. For quantitative comparison in the photorealism domain, we used the official implementation of Deep Image Blending (DIB)\footnote {\scriptsize \url{https://github.com/owenzlz/DeepImageBlending}} \cite{zhang2020deep}, Blended Diffusion\footnote {\scriptsize \url{https://github.com/omriav/blended-latent-diffusion}} \cite{avrahami2022blended_latent}, Paint by Example\footnote {\scriptsize \url{https://github.com/Fantasy-Studio/Paint-by-Example}} \cite{yang2022paint}, and SDEdit\footnote {\scriptsize \url{https://github.com/ermongroup/SDEdit}} \cite{meng2021sdedit}. Our framework utilizes Stable Diffusion\footnote {\scriptsize \url{https://github.com/Stability-AI/stablediffusion}} with the second-order DPM-Solver++ to solve all three ODEs in 20 steps. The first two inversion ODEs, aimed at obtaining accurate inverted noises and self-attention maps, were performed under the exceptional prompt with a common token value of 7788, while the last ODE utilized the normal prompt with a CFG scale of 2.5. The threshold values $\tau_A$ and $\tau_B$ were set at 0.4 and 0, respectively. 

\section{Ablation of Value Injection}
We conducted an additional ablation study in which we not only injected the attention maps but also included the values information. Specifically, we multiply the attention maps with the corresponding values for both the main and reference images, and then compose and inject them. The metrics obtained on the dataset are as follows: $\text{LPIPS}_\text{(BG)}=0.10$, $\text{LPIPS}_\text{(FG)}=0.63$, $\text{CLIP}_\text{(Image)}=81.37$, $\text{CLIP}_\text{(Text)}=27.68$. These metrics are lower compared to injecting only the attention maps. 

\begin{figure}[tbp]
	\centering
	\includegraphics[width=1\linewidth]{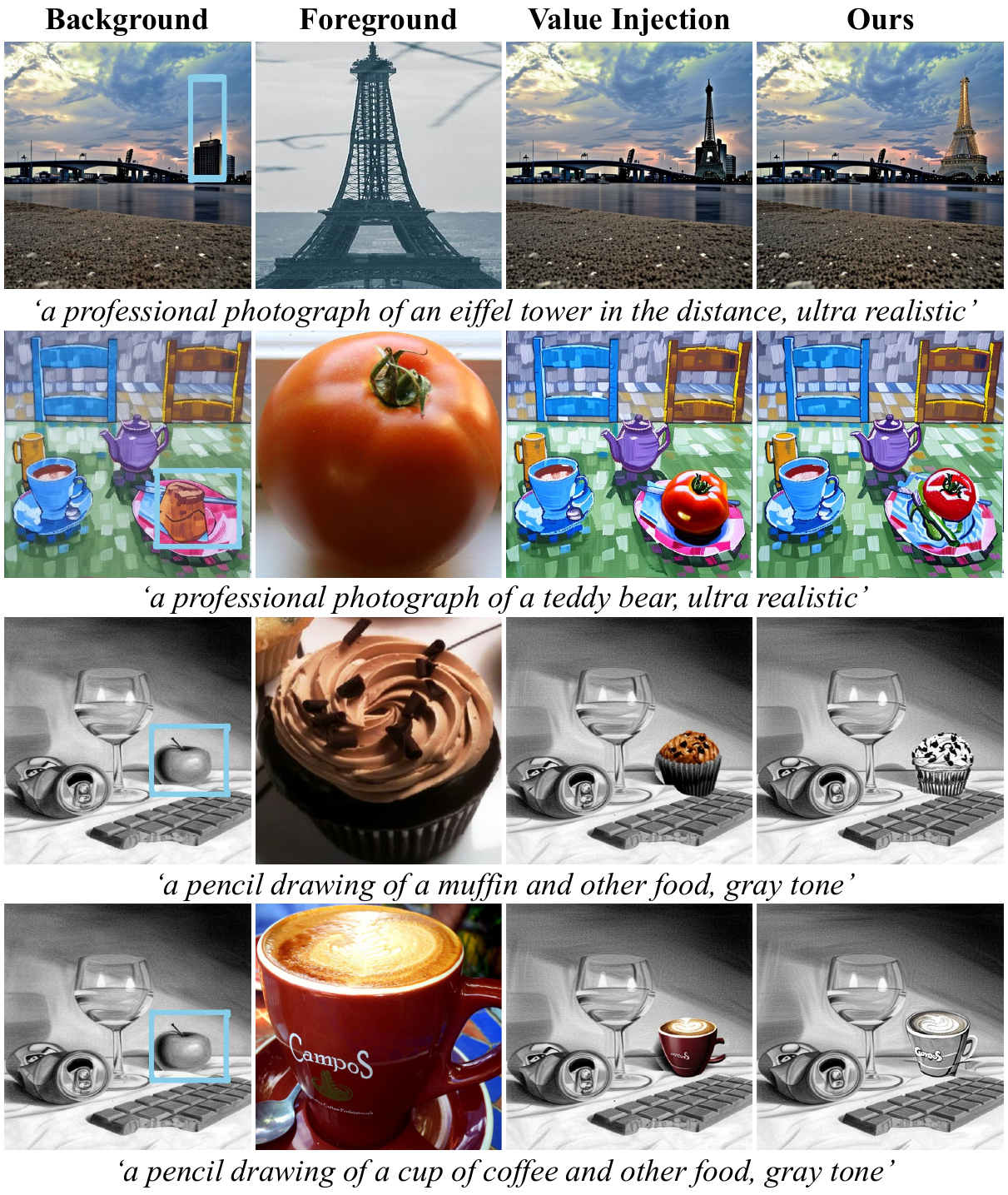}
	\vspace{-0.5cm}
	\caption{The visual comparison between injecting all information and our implementation. Injecting values leads to a more rigid generation, potentially impeding the smooth transition across visual domains. This impact becomes particularly evident when transferring to the sketchy domain.}
	\vspace{-0.3cm}
	\label{fig:inject_v}
\end{figure}

The rationale behind this is that injecting all the information might result in a more rigid generation, potentially hindering the ability to transition across visual domains due to the direct replacement of all information from the guiding images. On the other hand, by injecting self-attention maps only, we are able to preserve the semantic layouts while incorporating values derived from the inherent composition features. The visual comparison is shown in Figure~\ref{fig:inject_v}.

\section{User Study}
\label{user}
To compare image composition baselines across various domains, we conducted a user study by recruiting 50 participants from Amazon. The participants were asked to complete 40 ranking questions, with each question comprising a foreground image, a background image with a bounding box to indicate the region of interest, and a text prompt. For each question, the participants were presented with five images generated using different methods. They were requested to rank five images from 1 to 5 (1 being the best and 5 being the worst) based on comprehensive  criteria:
\begin{itemize}
	\item [1.] \textbf{Text Alignment:} The resulting image should match the specific style mentioned in the text prompt. For example, if the target domain is cartoon, oil painting, pencil drawing, or photorealism, the generated image should align with that style. 
	\item [2.] \textbf{Foreground Preservation:} The generated image should well-preserve the features or identity of the given object within the mask region, such that the viewers can recognize that the given and the generated objects are the same even in different domains.
	\item [3.] \textbf{Background Preservation:} The background outside the mask should remain unchanged.
	\item [4.] \textbf{Seamless Composition:} The resulting image should be of high quality and free from any apparent artifacts that might indicate it was generated by AI or copied and pasted.
\end{itemize}

\begin{figure}[tbp]
	\centering
	\includegraphics[width=1\linewidth]{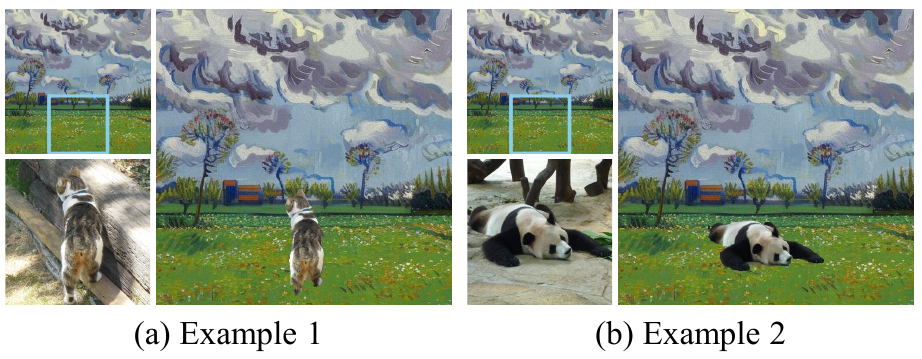}
	\caption{Examples of meaningless questions. The resulting images were generated by simply segmenting objects from the reference image and pasting them onto the region of interest in the background image without modifications. Despite the lack of any modification, the results appear almost seamless.}
	\label{fig:filter_question}
\end{figure}

To ensure all 40 questions are meaningful, we filtered out simple questions that, without any domain or illumination adjustment, only require copy-pasting operations to make the composition look natural despite the foreground and background being from different domains. We show examples of such cases in Figure~\ref{fig:filter_question}. After the filtering process, we randomly sampled questions from the test benchmark. In addition to the regular ranking questions, we also included three attention-checking questions to filter out random or invalid responses. The final valid questions consisted of 20 photorealism, 7 oil painting, 7 pencil sketching, and 6 cartoon animation questions.

The ranking score of the options in each question is calculated by: 
\begin{align}
	\text{score} = \frac{1}{n} \cdot \sum\limits_{i = 1}^5 {{v_i} \cdot {w_i}}
	\label{eqn:ranking}
\end{align}
where $v_i$ denotes the number of votes for the option to rank $i$, $w_i$ indicates the weight of rank $i$, and $n$ is the number of respondents. The first rank has the highest weight of 5 and the last rank has the lowest weight of 1. The resulting score reflects the overall ranking of the options, with a higher score indicating a better ranking.

\section{Self-Attention Visualization}
\label{sec:sa_visual}
\begin{figure}[tbp]
	\centering
	\includegraphics[width=1\linewidth]{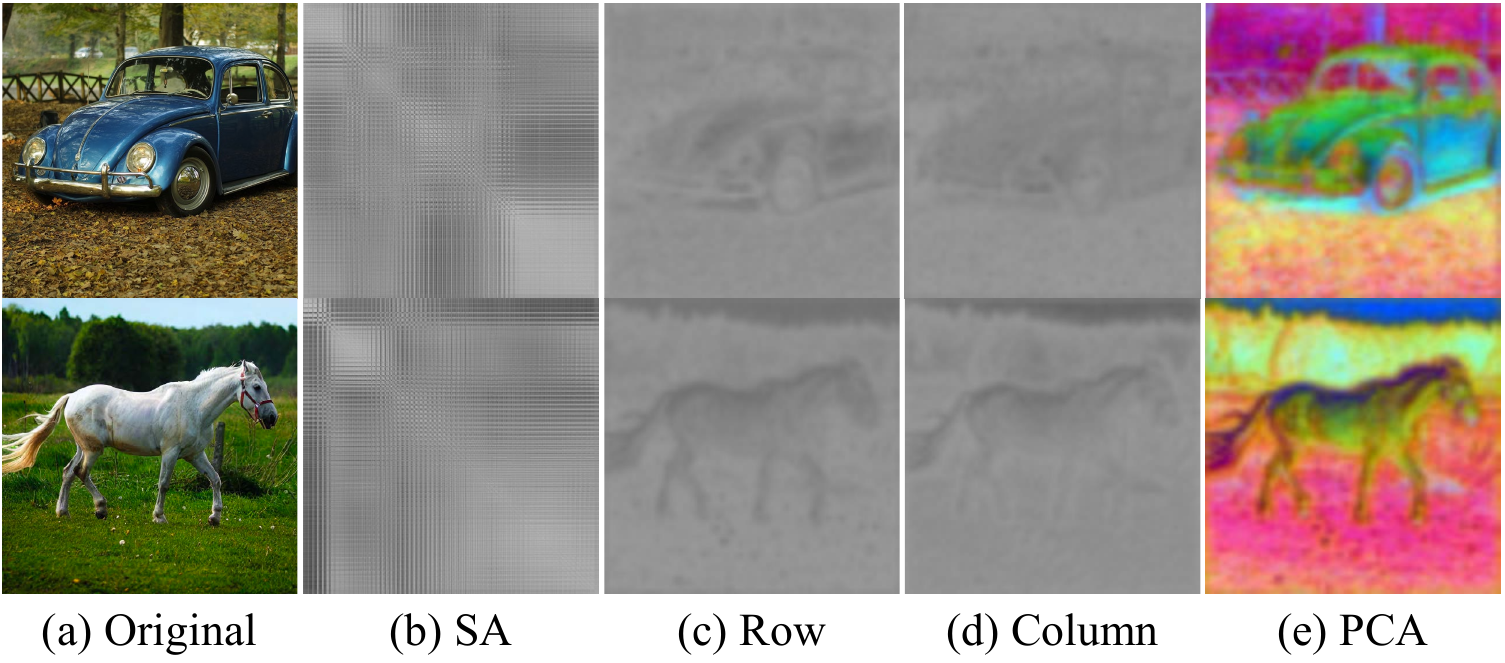}
	\caption{The visualization of (a) original image; (b) self-attention (SA) maps $\in {\mathbb{R}^{4096 \times 4096}}$ of (a); (c) the averaging result of unfolding all rows $ \in {\mathbb{R}^{1 \times 4096}}$ of (b) into ${\mathbb{R}^{64 \times 64}}$; (d) same operation as (c) for columns; (e) visualizing top-3 PCA components of (b).}
	\label{fig:sa_visual}
\end{figure}

Figure~\ref{fig:sa_visual} demonstrates how self-attention maps preserve semantic information. By unfolding the rows or columns of the self-attention maps, we can discern the underlying semantics of the image.

\section{Elaboration for Toy Example}
This section further analyzes the attention composition in Figure~\ref{fig:atten_inject}. The self-attention maps of the blue region in Figure~\ref{fig:atten_inject}~(a) $\bfA^\text{r}_{l,t}$ $\in {\mathbb{R}^{4 \times 4}}$ are partitioned into four blocks based on the patch indices and composed into the blue regions in $\bfA^\text{m}_{l,t}$ $\in {\mathbb{R}^{16 \times 16}}$, as illustrated in Figure~\ref{fig:atten_inject}~(b). The dimension of $\bfA^\text{cross}_{l,t}$ $\in {\mathbb{R}^{16 \times 4}}$ is identical to that of the green regions, with the exception of the interactions between white patches indexed at 5, 6, 9, and 10, and blue patches with corresponding indices. Since the aim of the attention composition is to infuse contextual information from the white region into the blue region, the information from the white patches indexed at 5, 6, 9, and 10 is irrelevant and can be disregarded.

\section{Test Benchmark}
\label{benchmark}
To facilitate evaluating cross-domain image-guided composition as a unified task, we have created a comprehensive test benchmark comprising 332 samples. Each sample in the benchmark comprises a main (background) image, a reference (foreground) image, a user mask, and a text prompt. Images were collected from Open Images \cite{kuznetsova2020open}, PASCAL VOC \cite{everingham2009pascal}, COCO \cite{lin2014microsoft}, Unsplash\footnote {\scriptsize \url{https://unsplash.com/}}, and Pinterest\footnote {\scriptsize \url{https://www.pinterest.com/}}. The main images comprise four visual domains: photorealism, pencil sketching, oil painting, and cartoon animation. All reference images are from the photorealism domain, as the reference requires segmentation models, which are generally more effective in this domain. The selection objective is to ensure that the main image and reference image share similar semantics, thereby guaranteeing a reasonable combination. The text prompt is manually labeled according to the semantics of the main and reference images. 

The reference images comprise a wide range of object classes, including `Car', `Panda', `Dog', `Elephant', `Fox', `Castle', `Buddha', `Bird', `Sheep', `Fire Hydrant', `Mailbox', `Hamburger', `Chicken', `Skyscraper', `Rocket', `Chair', `Cabinet', `Bag', `Teddy Bear', `Mall', `Tower', `Building', `Flower', `Tortoise', `Sparrow', `Ostrich', `Horse', `Cat', `Goose', `Tiger', `Eagle', `Squirrel', `Raccoon', `Penguin', `Sea Lion', `Goat', `Owl', `Microwave', `Bread', `Cake', `Tomato', `Fish', `Croissant', `Hot Dog', `Waffle', `Pancake', `Popcorn', `Burrito', `Muffin', `Juice', `Coffee', `Paper Towel', 'Tart', `Sandwich', `Teapot', `Lemon', `Candle', `Spoon', `Grapefruit', `Turkey', `Pomegranate', `Doughnut', `Cantaloupe', `Sandwich', `Cantaloupe', and `Turkey'. Given that most image composition baselines are trained exclusively on photorealistic images, our test benchmark contains a greater proportion of photorealism samples to enable a quantitative comparison. Specifically, the benchmark includes 237 photorealism samples, as well as 37 oil painting, 31 pencil sketching, and 27 cartoon animation samples. The benchmark will be publicly available for use in evaluating the performance of cross-domain image-guided composition methods.

\section{Additional Qualitative Results}
\subsection{Image Reconstruction}
\begin{figure*}[tbp]
	\centering
	\vspace{-0.2cm}
	\includegraphics[width=1\linewidth]{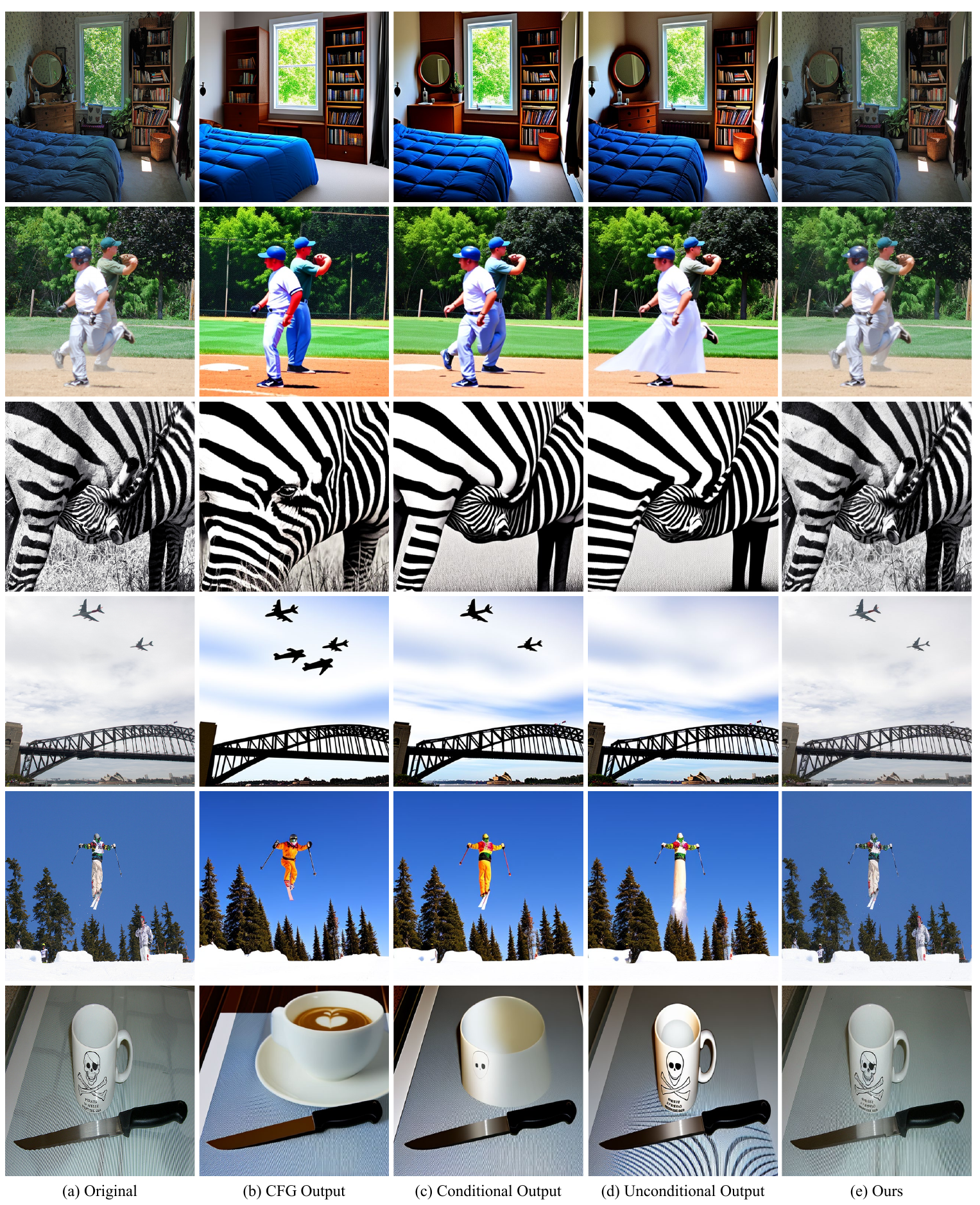}
	\vspace{-0.7cm}
	\caption{Comparison of image reconstruction results on the COCO using Stable Diffusion with (b) classifier-free guidance (CFG) output $\hat{\epsilon}_\theta(\bfx_t, t, \mathcal{E}, \varnothing)$, (c) conditional output $\epsilon_\theta(\bfx_t, t, \mathcal{E})$, (d) unconditional output $\epsilon_\theta(\bfx_t, t, \varnothing)$, and (e) ours $\epsilon_\theta(\bfx_t, t, \mathcal{W})$.}
	\label{fig:addi_coco}
\end{figure*}

\begin{figure*}[tbp]
	\centering
	\vspace{-0.2cm}
	\includegraphics[width=1\linewidth]{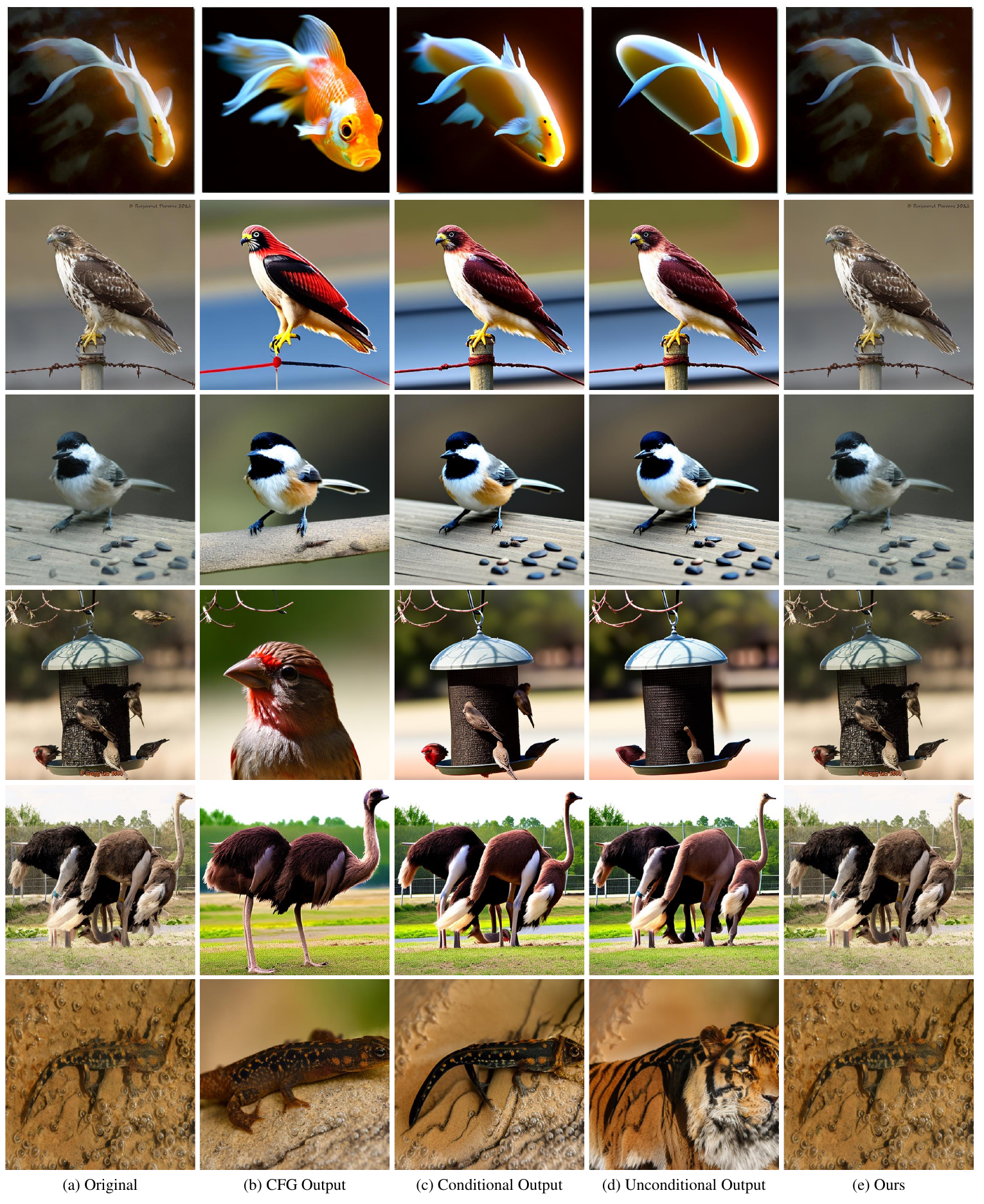}
	\vspace{-0.7cm}
	\caption{Comparison of image reconstruction results on the ImageNet using Stable Diffusion with (b) classifier-free guidance (CFG) output $\hat{\epsilon}_\theta(\bfx_t, t, \mathcal{E}, \varnothing)$, (c) conditional output $\epsilon_\theta(\bfx_t, t, \mathcal{E})$, (d) unconditional output $\epsilon_\theta(\bfx_t, t, \varnothing)$, and (e) ours $\epsilon_\theta(\bfx_t, t, \mathcal{W})$.}
	\label{fig:addi_imagenet}
\end{figure*}

\begin{figure*}[tbp]
	\centering
	\vspace{-0.2cm}
	\includegraphics[width=1\linewidth]{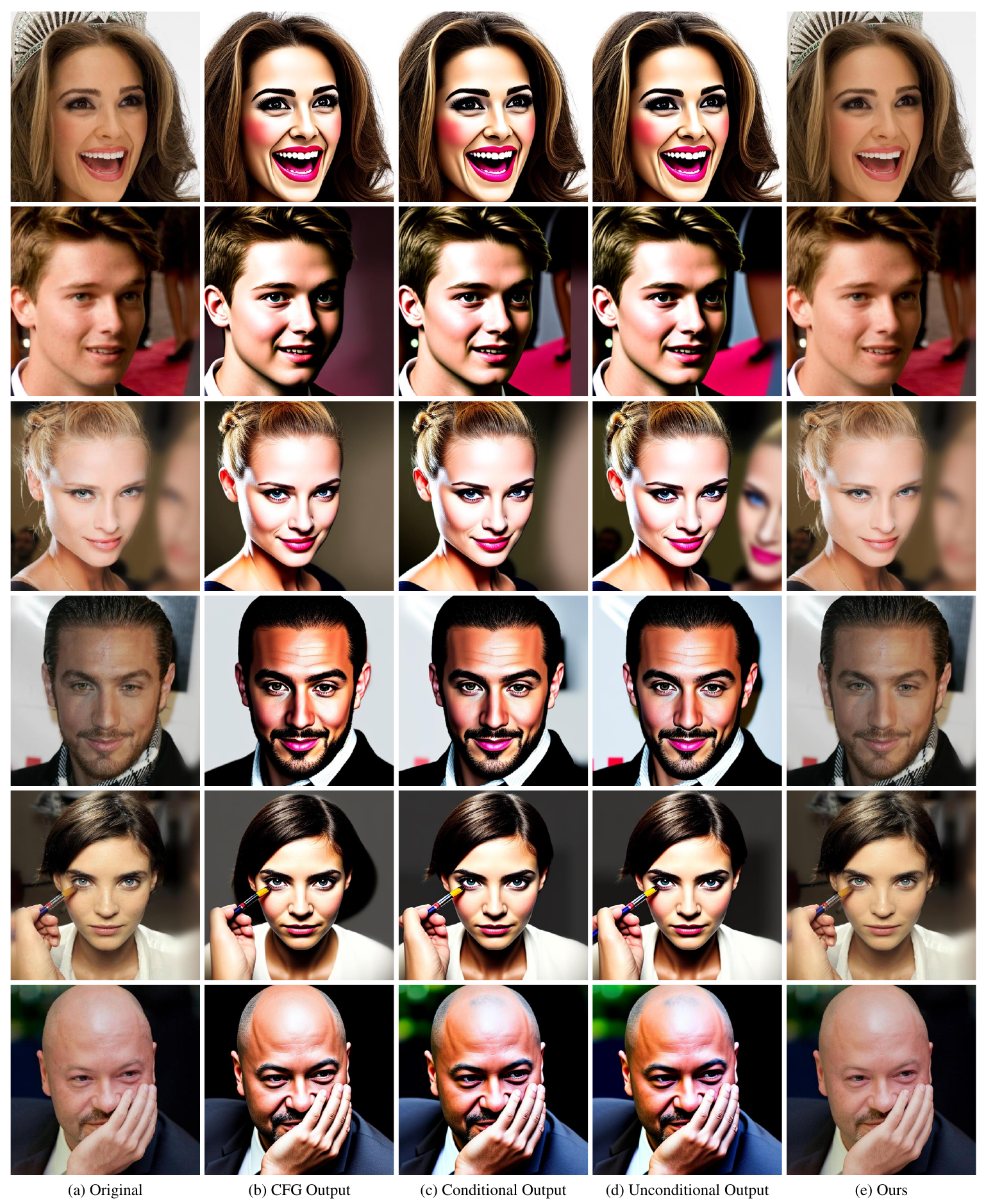}
	\vspace{-0.7cm}
	\caption{Comparison of image reconstruction results on the CelebA-HQ using Stable Diffusion with (b) classifier-free guidance (CFG) output $\hat{\epsilon}_\theta(\bfx_t, t, \mathcal{E}, \varnothing)$, (c) conditional output $\epsilon_\theta(\bfx_t, t, \mathcal{E})$, (d) unconditional output $\epsilon_\theta(\bfx_t, t, \varnothing)$, and (e) ours $\epsilon_\theta(\bfx_t, t, \mathcal{W})$.}
	\label{fig:addi_celeba}
\end{figure*}

Figures \ref{fig:addi_coco}, \ref{fig:addi_imagenet}, and \ref{fig:addi_celeba} present additional qualitative image reconstruction comparisons among different outputs of Stable Diffusion on COCO, ImageNet, and CelebA-HQ, respectively.
\subsection{Image Composition}
\label{sec:addi_qual_results}
\begin{figure*}[tbp]
	\centering
	\vspace{-0.2cm}
	\includegraphics[width=1\linewidth]{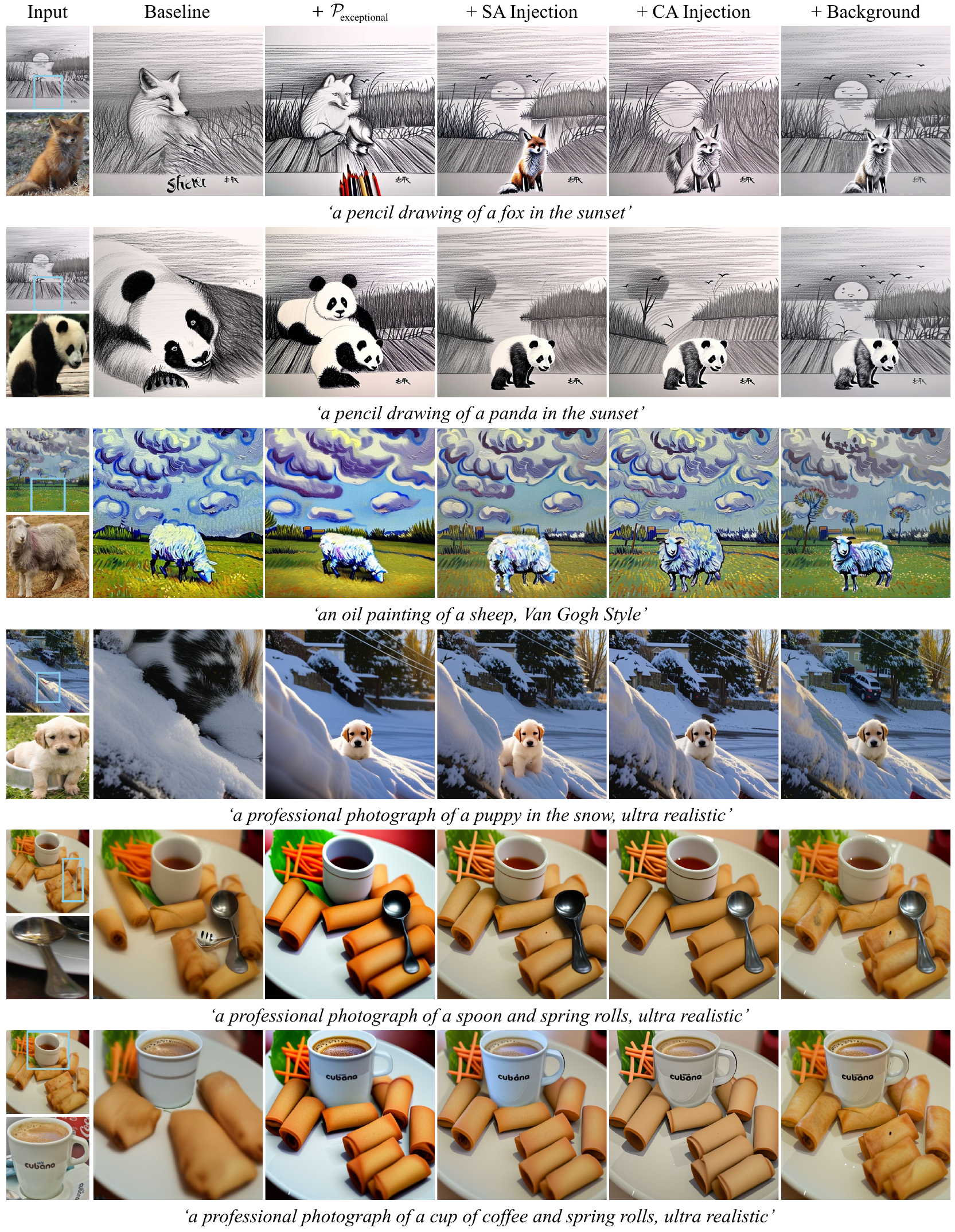}
	\vspace{-0.8cm}
	\caption{Ablation study of different variants of our framework. SA: self-attention. CA: cross-attention.}
	\label{fig:ablation2}
\end{figure*}

\begin{figure*}[tbp]
	\centering
	\vspace{-0.3cm}
	\includegraphics[width=1\linewidth]{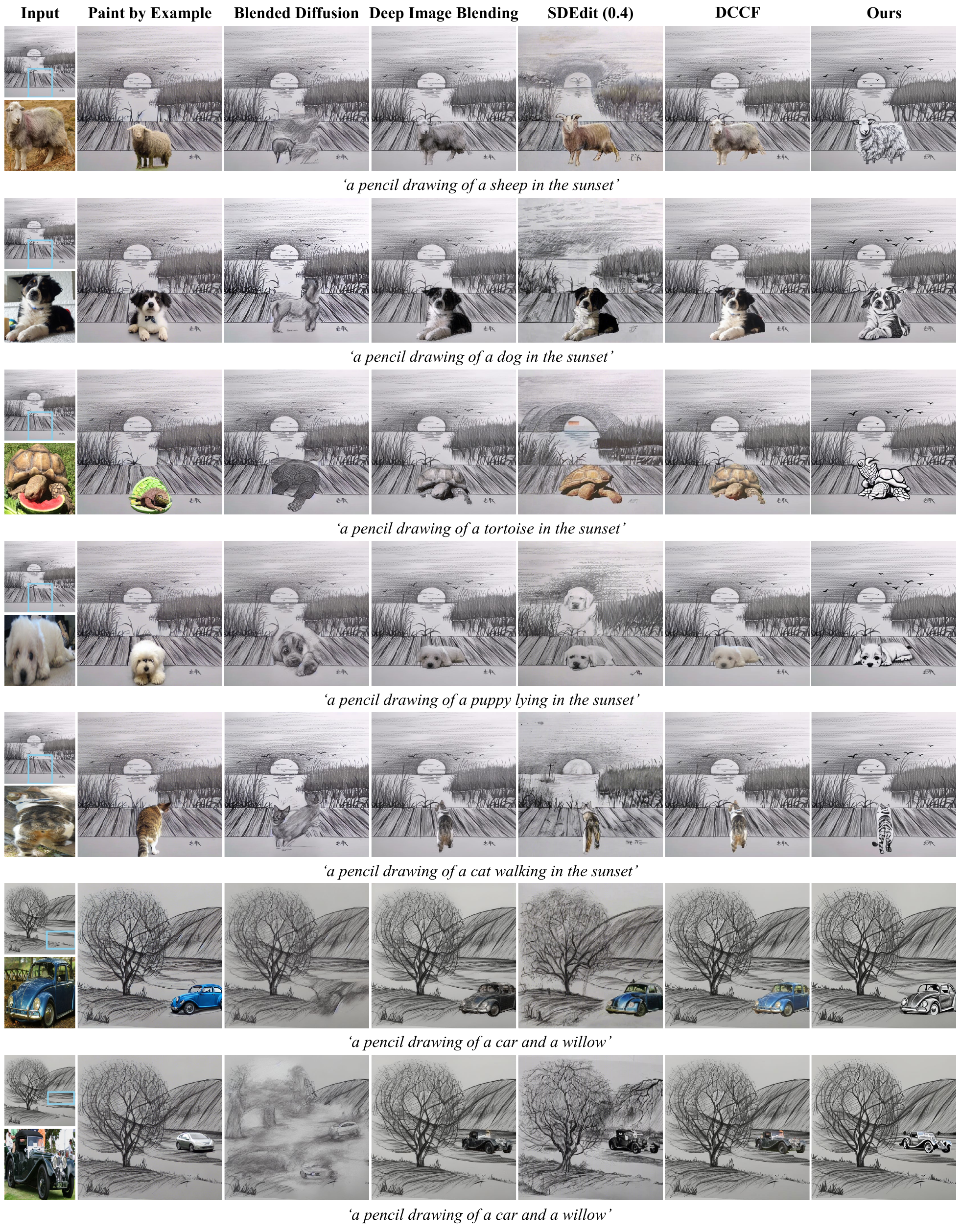}
	\vspace{-0.8cm}
	\caption{Qualitative comparison with SOTA baselines in image composition for the pencil sketching domain.}
	\label{fig:addi_com_ske}
\end{figure*}

\begin{figure*}[tbp]
	\centering
	\vspace{-0.3cm}
	\includegraphics[width=1\linewidth]{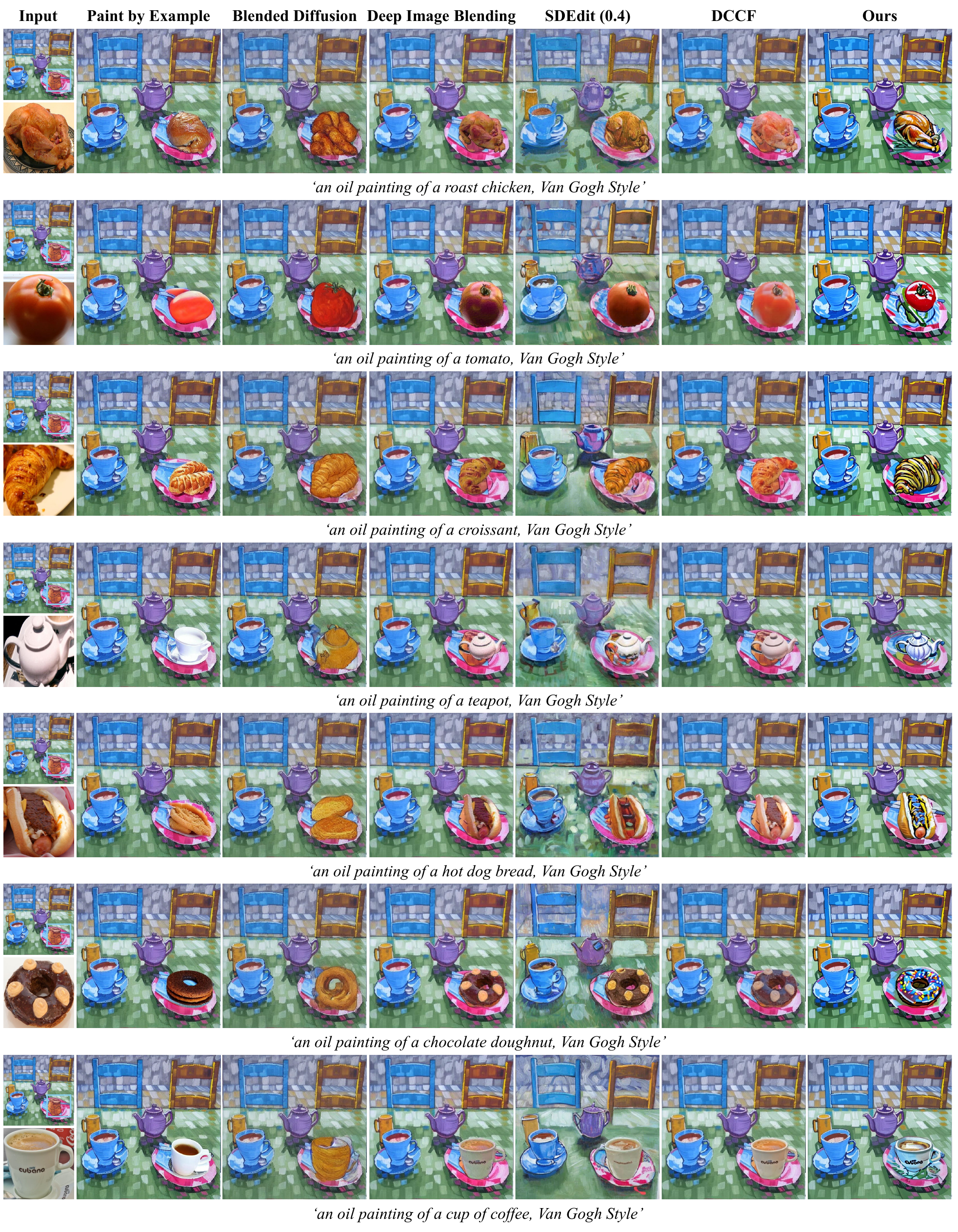}
	\vspace{-0.8cm}
	\caption{Qualitative comparison with SOTA baselines in image composition for the oil painting domain.}
	\label{fig:addi_com_oil}
\end{figure*}

\begin{figure*}[tbp]
	\centering
	\vspace{-0.3cm}
	\includegraphics[width=1\linewidth]{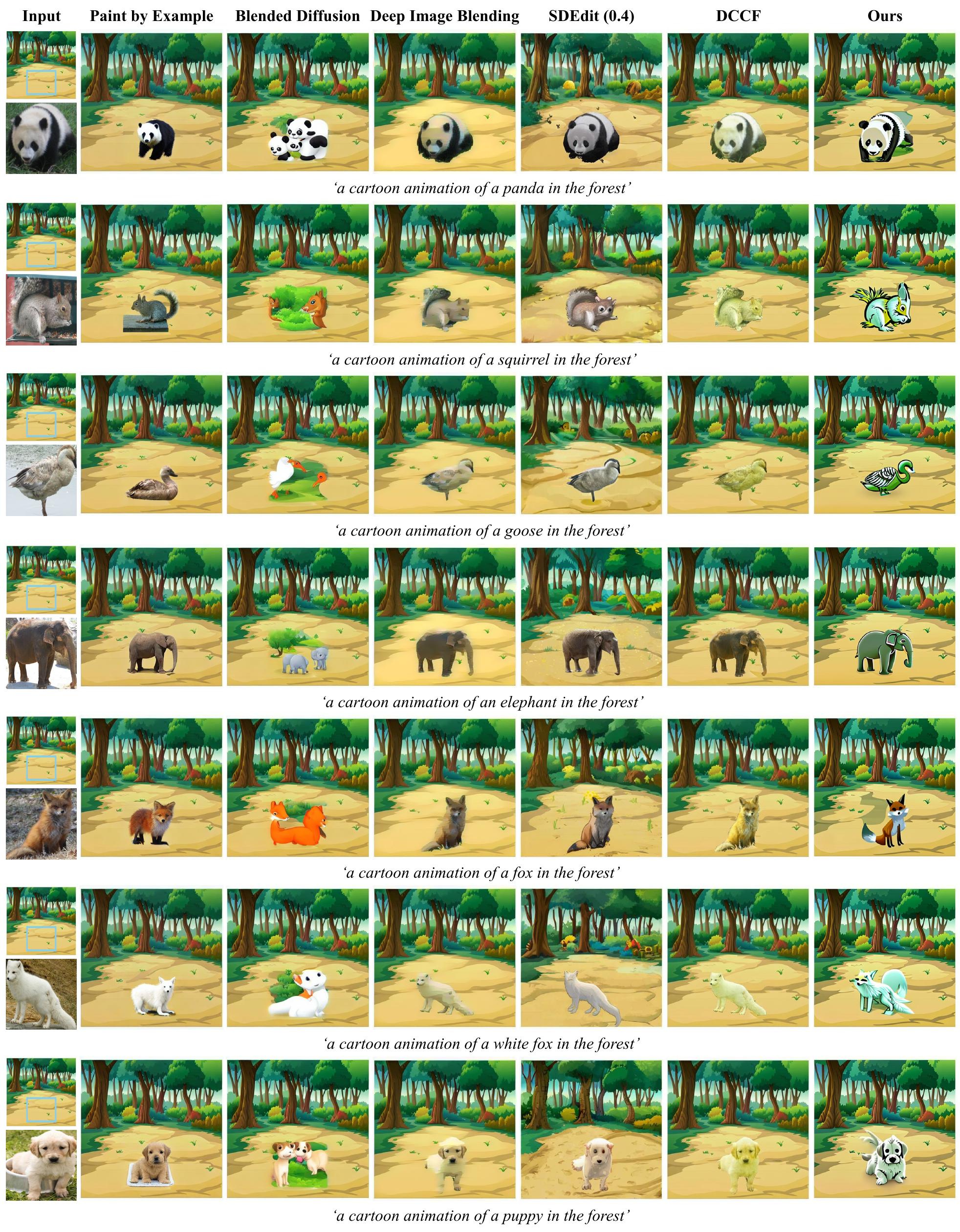}
	\vspace{-0.8cm}
	\caption{Qualitative comparison with SOTA baselines in image composition for the cartoon animation domain.}
	\label{fig:addi_com_carton}
\end{figure*}

\begin{figure*}[tbp]
	\centering
	\vspace{-0.3cm}
	\includegraphics[width=1\linewidth]{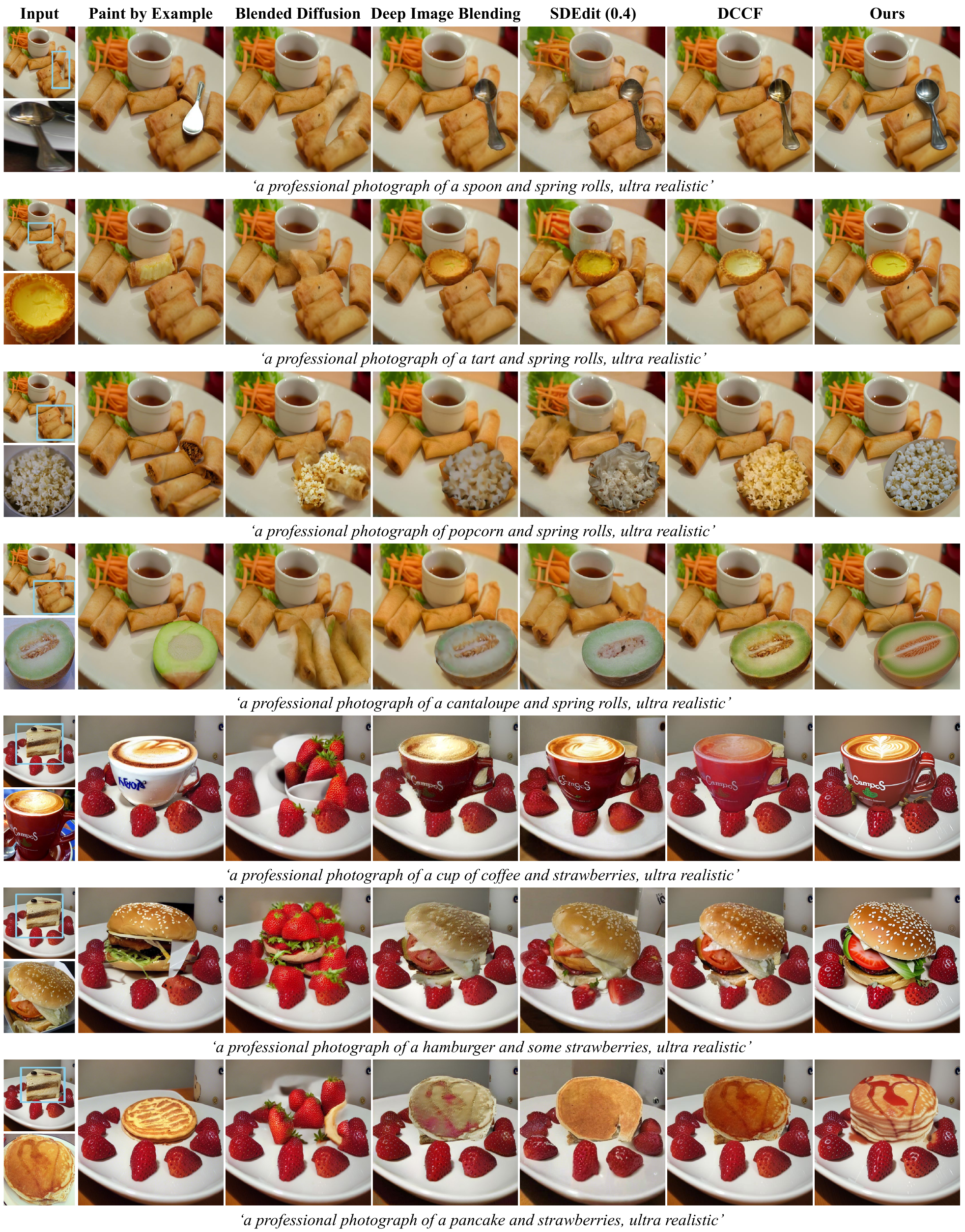}
	\vspace{-0.7cm}
	\caption{Qualitative comparison with SOTA baselines in image composition for the photorealism domain.}
	\label{fig:addi_com_real1}
\end{figure*}

\begin{figure*}[tbp]
	\centering
	\vspace{-0.3cm}
	\includegraphics[width=1\linewidth]{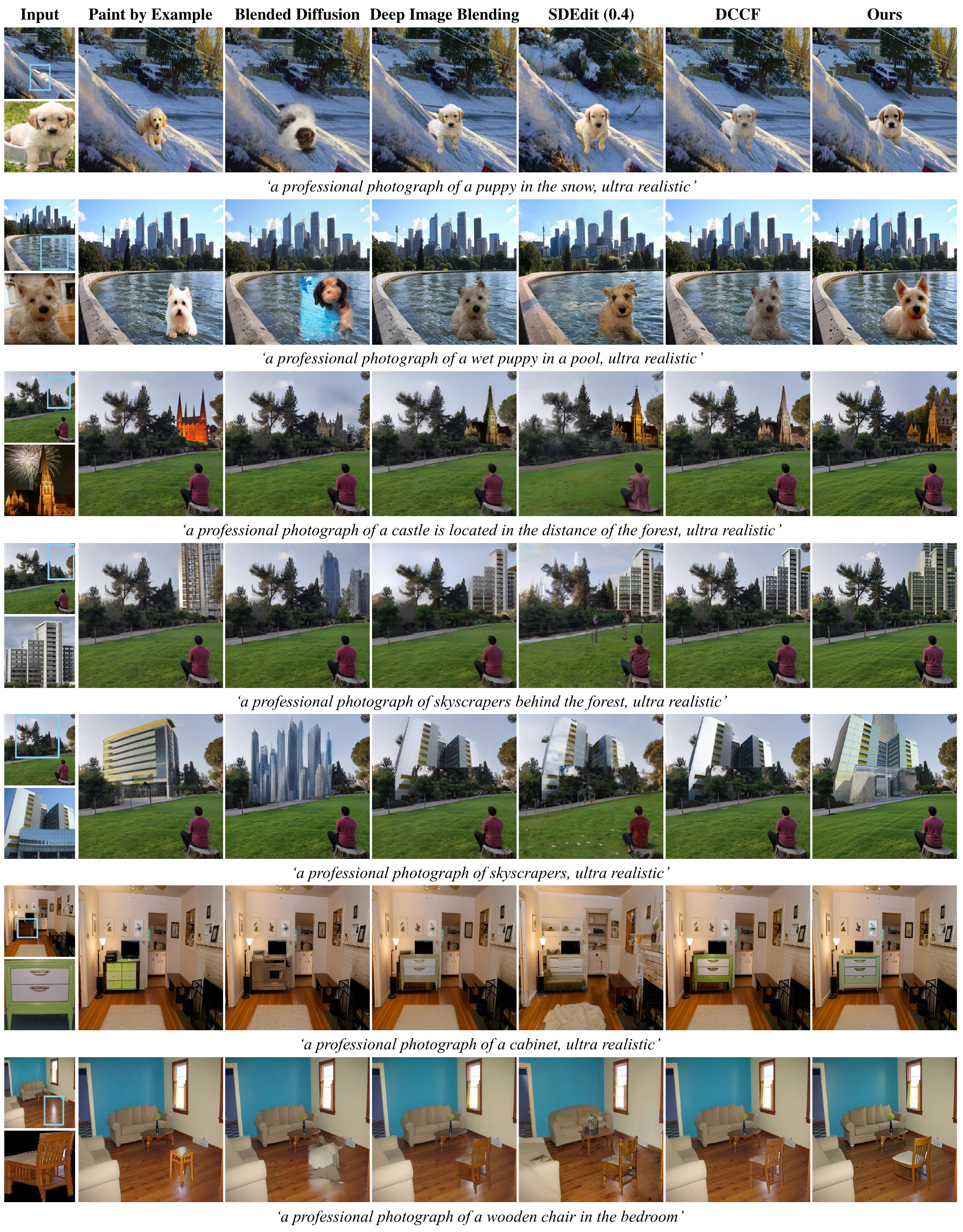}
	\vspace{-0.7cm}
	\caption{Qualitative comparison with SOTA baselines in image composition for the photorealism domain.}
	\label{fig:addi_com_real2}
\end{figure*}

\begin{figure*}[tbp]
	\centering
	\vspace{-0.3cm}
	\includegraphics[width=1\linewidth]{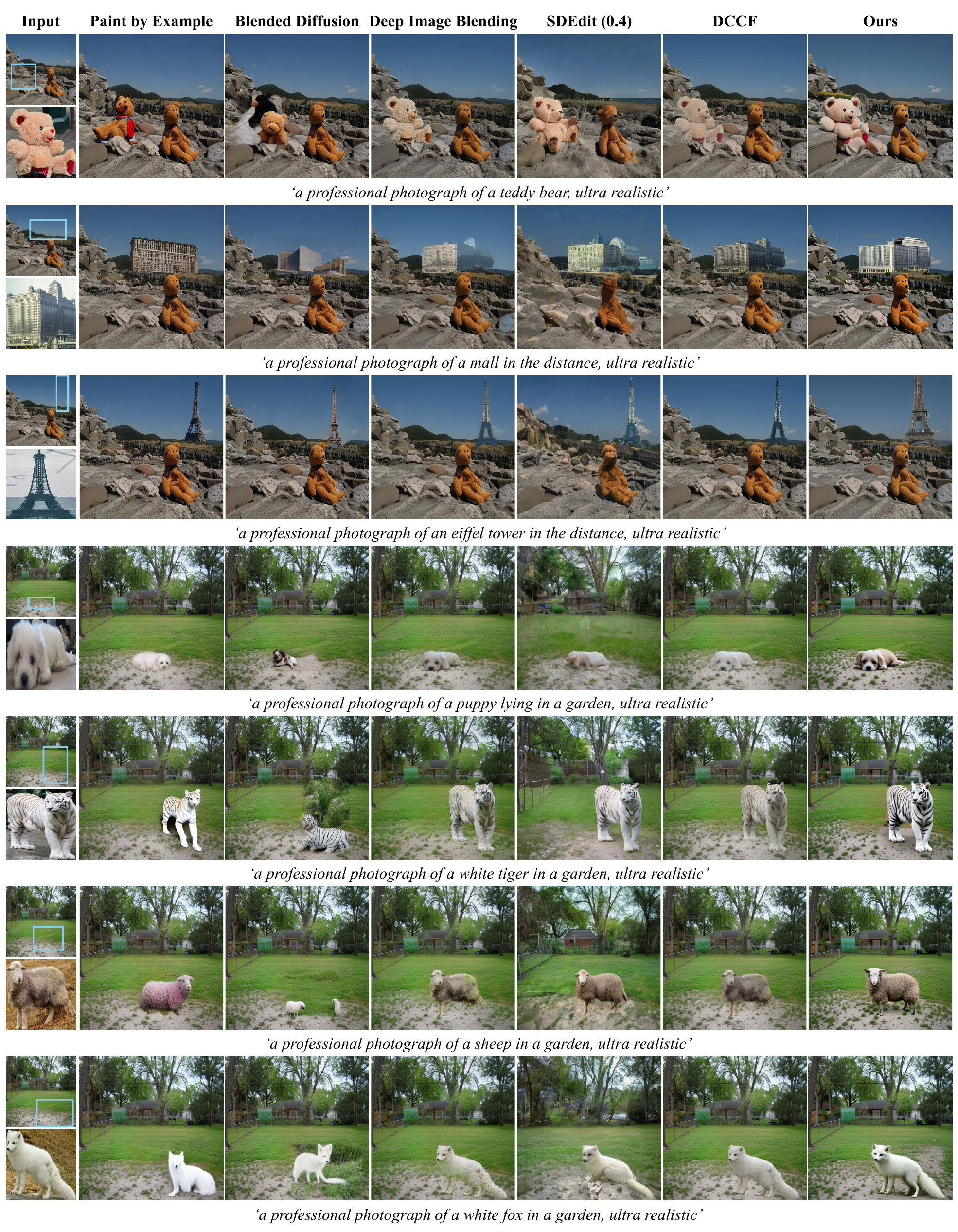}
	\vspace{-0.7cm}
	\caption{Qualitative comparison with SOTA baselines in image composition for the photorealism domain.}
	\label{fig:addi_com_real3}
\end{figure*}

Figure~\ref{fig:ablation2} presents additional ablation study results. Further qualitative comparisons of image composition across various domains are exhibited in Figures \ref{fig:addi_com_ske}, \ref{fig:addi_com_oil}, \ref{fig:addi_com_carton}, \ref{fig:addi_com_real1}, \ref{fig:addi_com_real2}, and \ref{fig:addi_com_real3}.

\section{Societal Impacts}
TF-ICON offers a means of image-guided composition that empowers individuals without professional artistic skills to create compositions. While this technology is beneficial, it can also be misused for malicious purposes, such as in cases of harassment or spreading fake news. Moreover, image composition is closely related to image generation, so it is essential to recognize that using diffusion models trained on web-scraped data, such as LAION \cite{schuhmann2022laion}, can potentially introduce biases. Specifically, LAION has been found to contain inappropriate content such as violence, hate, and pornography, as well as racial and gender stereotypes. Consequently, diffusion models trained on LAION, such as Stable Diffusion and Imagen \cite{saharia2022photorealistic}, are prone to exhibit social and cultural biases. As such, using such models raises ethical concerns and should be approached with care. Finally, the capacity to compose across artistic domains has the potential to be exploited for copyright infringement purposes, as users could generate images in a similar style without the consent of the artist. Although the resulting generated artwork may be readily distinguishable from the original, future technological advances could render such infringement challenging to identify or legally prosecute. Thus, we encourage users to use this method cautiously and only for appropriate purposes.

\end{document}